\documentclass[12pt,a4paper]{article}
\usepackage[english]{babel}
\usepackage[letterpaper,top=2cm,bottom=2cm,left=3cm,right=3cm,marginparwidth=1.75cm]{geometry}
\usepackage[numbers]{natbib}
\usepackage{graphicx}
\usepackage[utf8]{inputenc}
\usepackage{amsmath}
\usepackage{amssymb}
\usepackage{tikz}
\usetikzlibrary{calc,trees,positioning,arrows,chains,shapes.geometric,%
    decorations.pathreplacing,decorations.pathmorphing,shapes,%
    matrix,shapes.symbols,shadows}
\usepackage[font={small,it}]{caption}
\usepackage{subcaption}
\usepackage[colorlinks=true, allcolors=blue]{hyperref}
\usepackage{comment}
\usepackage{algorithm}
\usepackage{algpseudocode}
\usepackage{authblk}
\usepackage{indentfirst}
\usepackage{lineno, hyperref}
\modulolinenumbers[5]

\begin{document}

\title{Entropy-Enhanced Conformal Features from Ricci Flow for Robust Alzheimer’s Disease Classification}

\author[1]{Fatemeh Ahmadi\thanks{Corresponding author. Email: f.ahmadi@aut.ac.ir}}
\author[1]{Behroz Bidabad}	
\author[2,3]{Hamid Nasiri}
\affil[1]{\small Department of Mathematics and Computer Science, Amirkabir University of Technology (Tehran Polytechnic), Tehran, Iran.}	
\affil[2]{Department of Computer Engineering, Amirkabir University of Technology (Tehran Polytechnic), Tehran, Iran.}
\affil[3]{School of Computing and Communications, Lancaster University, Lancaster, UK.}
\date{}
\maketitle

\abstract{\noindent\textbf{Background and Objective:} In brain imaging, geometric surface models are essential for analyzing the 3D shapes of anatomical structures. Alzheimer's disease (AD) is associated with significant cortical atrophy, making such shape analysis a valuable diagnostic tool. The objective of this study is to introduce and validate a novel local surface representation method for the automated and accurate diagnosis of AD.\\
\textbf{Methods:} The study utilizes T1-weighted MRI scans from 160 participants (80 AD patients and 80 healthy controls) from the Alzheimer’s Disease Neuroimaging Initiative (ADNI). Cortical surface models were reconstructed from the MRI data using Freesurfer. Key geometric attributes were computed from the 3D meshes. \textit{Area distortion} and conformal factor were derived using Ricci flow for conformal parameterization, while Gaussian curvature was calculated directly from the mesh geometry. Shannon entropy was applied to these three features to create compact and informative feature vectors. The feature vectors were used to train and evaluate a suite of classifiers (e.g. XGBoost, MLP, Logistic Regression, etc.).\\
\textbf{Results:} Statistical significance of performance differences between classifiers was evaluated using paired Welch’s t-test. The method proved highly effective in distinguishing AD patients from healthy controls. The Multi-Layer Perceptron (MLP) and Logistic Regression classifiers outperformed all others, achieving an accuracy and F$_1$ Score of 98.62\%.\\
\textbf{Conclusions:} This study confirms that the entropy of conformally-derived geometric features provides a powerful and robust metric for cortical morphometry. The high classification accuracy underscores the method's potential to enhance the study and diagnosis of Alzheimer's disease, offering a straightforward yet powerful tool for clinical research applications.}



\maketitle

\section{Introduction}\label{sec1}
Surface-based models are essential in brain imaging, serving as tools to examine anatomical structures, detect abnormalities of the brain surface, and compare three-dimensional anatomical shapes between individuals. These models provide critical insights into brain morphology and its variations. Determining an exact bijective mapping between surfaces presents significant challenges, though techniques like surface registration and parameterization offer potential solutions \cite{zitova2003image}.

Surface parameterization is a method that parameterizes the surface and embeds it in standard parameter spaces. In recent years, many surface parameterization techniques have been introduced. Certain techniques directly apply parameterization and incorporate the cortical surface into the sphere domain by optimizing specific energy functions \cite{yueh2017efficient}. Alternative techniques do a planar parameterization that embeds the surface in a plane \cite{gravesen2014planar}.

In the surface parameterization, conformal methods place greater limitations on the surface's shape compared to a topological structure, making them more stable than Riemannian metrics. Ricci flow is a parameterization technique that conformally transforms the surface into a standard parameter space, such as a Euclidean plane, sphere, or hyperbolic plane \cite{zeng2013ricci}. On both smooth and discrete surfaces, the Ricci flow adjusts a Riemannian metric conformally, producing a space with consistent Gaussian curvature. In the discrete framework, the Ricci flow is resolved with Ricci energy optimization using Newton's method or gradient descent, based on the circle packing metric \cite{Jin2018Discrete}.

Over the last ten years, a multitude of research studies have employed discrete Ricci flow in the morphometry of brain surfaces to detect brain irregularities and disorders. In \cite{zeng2013teichmuller}, the authors calculated a Teichmüller shape descriptor using Ricci flow, which encodes information about both the local and global contours of a closed surface with zero genus. This novel feature is used to examine irregularities in the cortical brain surface, enabling Alzheimer's disease to be detected. Chen et al. \cite{chen2013ricci} introduced an innovative approach for calculating spherical parameterization through the application of Euclidean Ricci flow, which modified the computation of Gaussian curvature. They have exploited Ricci's energy for scale-space processing, enabling the extraction of scale-dependent geometric feature points essential for matching and registering surfaces. They applied Ricci energy to evaluate the shape of the hippocampus in individuals diagnosed with Schizophrenia. In \cite{su2015shape}, the authors present a novel framework for classifying brain cortical surfaces utilizing Wasserstein distance, grounded in the Ricci flow from the Riemannian optimal mass transport theory. This proposed method is employed to categorize the cortical surfaces of the brain based on different intelligence quotients. Shi et al. \cite{shi2017conformal} proposed a method to map multiple linked surfaces to the Poincaré disk using the surface Ricci flow technique. They calculated a collection of conformally invariant shape indices within the hyperbolic parameter domain, which correlate with the boundary lengths determined by the landmark curve. Their method was evaluated using 3D MRI data from ADNI to assess abnormalities in brain morphometry related to Alzheimer's disease (AD).

In \cite{shi2019hyperbolic}, the authors introduce a new framework for calculating the Wasserstein distance between generic surfaces using hyperbolic Ricci flow and harmonic mapping. This method was evaluated through studies on face recognition and monitoring the progression of Alzheimer's disease (AD). Khodaei et al. \cite{khodaei2024classification} used discrete Ricci flow as a spherical conformal parameterization method on genus-zero hippocampus surface without boundaries to Alzheimer's disease diagnosis. 

In addition to the application of Ricci flow to discrete surfaces, it has also been used in recent years in several graph-related machine learning applications, including community detection in complex networks \cite{karampour2025discrete, lai2022normalized}, high-dimensional data classification, dimension reduction and visualization \cite{xu2023camel}. They demonstrate the ability of Ricci flow in various applications.  

In disease diagnosis by analyzing cortical surfaces using Ricci flow, the methods frequently depend on specific landmarks to direct the analysis. While manual landmarking proves to be more effective than standard surface attribute processing, it poses significant challenges when applied to large datasets and necessitates specialized medical expertise. This can limit its scalability and accessibility in clinical practice, highlighting the need for automated or semi-automated methods to enhance the analysis of cortical surfaces in extensive studies. Therefore, in our current study, we concentrate on a method that eliminates the need for landmarks and focuses on processing significant local regions of the cerebral cortex that are affected by Alzheimer's disease. This approach aims to enhance the analysis and understanding of cortical changes associated with the disease without the limitations posed by manual landmarking. Since Alzheimer's disease primarily affects the hippocampal region, it is more efficient to analyze a specific region than the entire brain.\\
In one of our prior works \citep{AHMADI2024106212}, we proposed covariance-based descriptors to optimize Ricci energy on brain surfaces, modeled as 3D shapes. Since these descriptors belong to the nonlinear manifold of symmetric positive-definite matrices, we adopted Gaussian radial basis functions for manifold-based classification in the 3D shape analysis framework. This method was applied to study abnormal cortical morphometry for Alzheimer’s disease diagnosis. In our earlier research \citep{ahmadi2024alzheimer} that serves as the foundation for this study, employing Euclidean Ricci flow, we achieved planar parameterization and derived surface conformal representation for the hippocampal region. These features were then compressed using Shannon entropy. Subsequently, we used XGBoost, SVM, and Random Forest classifiers for Alzheimer’s disease detection.

The current study employs Euclidean Ricci flow for planar parameterization, extracting key geometric features including area distortion and conformal factor. Additionally, Gaussian curvature is computed directly on the surface mesh prior to parameterization. The extracted features are further encoded using Shannon entropy. For classification, we evaluate and compare the performance of multiple machine learning models: XGBoost, SVM, Random Forest, MLP, Decision Tree, KNN, and Logistic Regression. The proposed method has demonstrated its effectiveness in analyzing the cortical surface of the hippocampal region in both Alzheimer's disease patients and cognitively normal individuals. This application highlights its potential for distinguishing morphometric differences associated with cognitive decline. We conduct measurements specifically on the hippocampal region to assess structural changes associated with Alzheimer's disease. This targeted analysis allows us to identify morphometric variations that may correlate with cognitive impairments, ultimately aiding in the diagnosis and understanding of the disease's progression \citep{garg2023review}. 

Here's a summary of the paper's main contributions:
\begin{itemize}
\item A novel approach for binary classification between Alzheimer's disease and cognitively normal individuals using brain surface analysis has been proposed. 
\item Signatures involving area distortion, conformal factor are computed through Ricci flow parameterization, and Gaussian curvature is computed directly on the surface mesh prior to parameterization. These features on the surface are encoded using Shannon entropy.
\item In the classification phase, we use different classifiers such as eXtreme Gradient Boosting (XGBoost), Support Vector Machine (SVM), Random Forest, K-Nearest Neighbor (KNN), Multi-Layer Perceptron (MLP), Decision Tree, and Logistic Regression, and compare the results.
\end{itemize}

The structure of the paper is as follows: Section 2 provides the proposed method, it starts with mathematical foundation and preliminaries. Section 3 discusses the results of the classification. The discussion of the results is detailed in Section 4. Lastly, Section 5 offers the conclusions.

\section{Methods}
In this section, we present a novel approach to detecting the distinctive regions of a 3D mesh by applying the well-known concept of Shannon entropy to the values extracted from the Ricci flow on 3D mesh data. Our methodology comprises several key steps. First, we introduce the Ricci flow-based framework for surface parameterization. It is done using the definition of circle packing metric, Ricci energy optimization, and the plane embedding that was introduced in the following subsection. Next, we present the derivation of geometric signatures and their transformation into a compact representation via Shannon entropy. The pipeline concludes with the classification of subjects into Alzheimer's disease and Cognitively Normal groups.\\
\subsection{Ricci flow}
\subsubsection{Surface Ricci flow}
Computing the conformal representation of a surface could be a challenging issue. One of the capable specialized apparatuses for this work is Ricci flow. Hamilton introduced the Ricci flow as a tool to prove Poincaré's conjecture \cite{hamilton1982three}. The Ricci flow deforms the Riemannian metric in accordance with the Gaussian curvature, in a way that the curvature changes according to a nonlinear heat diffusion process, ultimately leading to a situation where the curvature becomes uniform throughout. Let $g_{ij}$ be a Riemannian metric and $K$ be the Gaussian curvature, the smooth Ricci flow introduces by:
\begin{equation} \label{eqConRicci}
    \frac{dg_{ij}(t)}{dt} = -2K(t)g_{ij}(t),
\end{equation}
where $t$ is the time factor.

It has been shown that the Ricci flow converges in a short time. Hamilton \cite{hamilton1982three} provided the proof for surfaces with non-positive Euler characteristic, while Chow \cite{chow1991ricci} addressed the case for surfaces with positive Euler characteristic. 
Geometrically, Ricci flow can be understood as a method of smoothing the metric. High curvature regions (peaks) will shrink, while low curvature regions will expand, effectively balancing the shape of the surface over time. This smoothing effect is beneficial in analyzing brain anatomy, as it helps highlight anatomical differences that correlate with Alzheimer's progression.

\subsubsection{Discrete Ricci flow}
Discrete Ricci flow generalizes the concept of Ricci flow to discrete surfaces (meshes). Triangular meshes are a standard method for approximating smooth surfaces and are crucial for processing the 3D data obtained from medical imaging. Consider a mesh $ M $ represented by the set of vertices $ V $ connected by edges $ E $. The metric is described by the lengths of edges, and the curvature is defined for each vertex based on angles incident to that vertex.

\textbf{Discrete curvature calculation:} For a vertex in a mesh, the discrete Gaussian curvature can be estimated using angles within the triangles formed by its neighboring vertices. The curvature gives insight into how the local geometry differs from flat geometry. The discrete Gaussian curvature is defined by:
\begin{equation} \label{eqGaussCurv}
    K(v_i)=\begin{cases}     
    \pi-\sum_{j}\theta_j &  for \; boundray \; vertices\\
    2\pi-\sum_{j}\theta_j & for \; interior \; vertices 
     \end{cases}
\end{equation}
where $ \theta_j $ are the angles at vertex $ v_i $ formed with its neighboring vertices.

\begin{figure*}[h!]
\centering
\begin{subfigure}{0.3\textwidth}
\includegraphics[width=\textwidth]{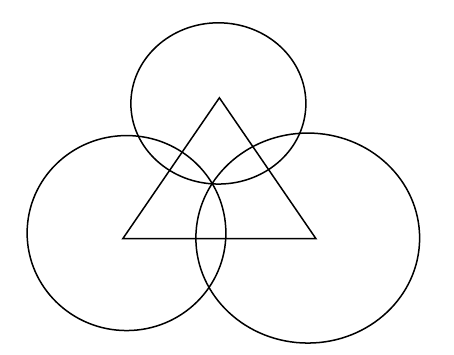}
\caption{}
\end{subfigure}
\begin{subfigure}{0.3\textwidth}
\includegraphics[width=\textwidth]{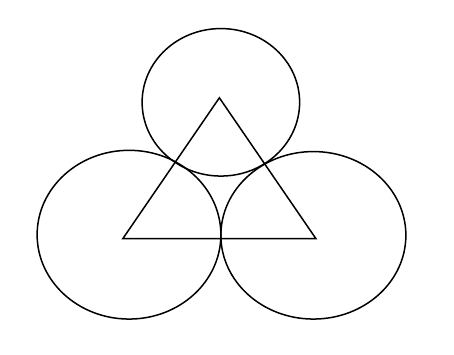}
\caption{}
\end{subfigure}
\begin{subfigure}{0.3\textwidth}
\includegraphics[width=\textwidth]{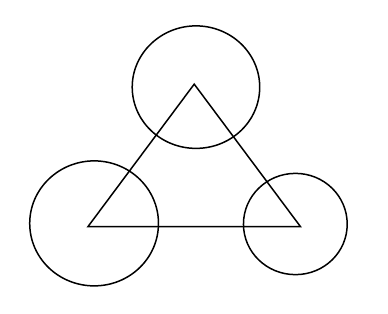}
\caption{}
\end{subfigure}
\caption{Circle packing schemes. (a): Thurston’s circle packing, (b): Tangential circle packing, (c): Inversive distance circle packing.}
\label{fig:CirclePacking}
\end{figure*}

The initial configuration for Ricci flow often starts with the circle packing metrics, which are a way of representing a discrete metric on a surface. Figure \ref{fig:CirclePacking} illustrates the general schemes of circle packing metrics. Depending on whether the circles are \textit{intersecting}, \textit{tangent} or \textit{disjoint}, there are three different patterns of circle packing, called \textit{Thurston}, \textit{tangential} and \textit{inversive distance circle packing}, respectively \cite{zhang2014unified}. 
\subsubsection{Ricci flow algorithm}
For computing a Riemannian metric that is conformal to the original metric and satisfy the target curvature, the Ricci flow is applied on discrete surfaces with an Euclidean background geometry \cite{Jin2018Discrete}. The algorithm of Ricci flow involving three steps: initial circle packing, Ricci energy optimization, and plane embedding is reviewed as follows.

\textbf{Initial circle packing:}
The initial configuration for Ricci flow often starts with circle packings, which is a way of representing a discrete metric on a surface. Given a collection of disjoint circles on a surface, the packing distances between the centers of these circles. \\
\textbf{Algorithm for inversive distance circle packing:} Suppose the triangular mesh $M$ with the initial Euclidean metric and the given target curvature $\bar{K}: V \rightarrow \mathbb{R}$, where $V$ is the vertex set of the mesh. For each vertex $ v_i$ surrounded by faces $f_{ijk}$, the circle's radius is computed for the vertex $ v_i$ using edge length as:
\begin{equation}
	\gamma_i^{jk} = \dfrac{l_{ij}+l_{ki} - l_{jk}}{2},
\end{equation}
where $ l_{ij} $ is the distance between the vertices $ v_i $ and $ v_j $.
For each edge $e_{ij} $, the inverse distance between the two circles $(v_i,\gamma_i)$ and $(v_j,\gamma_j)$ is defined by 
\begin{equation}
	\eta_{ij}=\dfrac{l_{ij}^2-\gamma_i^2-\gamma_j^2}{2\gamma_i\gamma_j}.
\end{equation}

The pipeline of the initial circle packing metric is represented in Algorithm \ref{algorithm1}.

\begin{algorithm}[!h] \fontsize{10pt}{10pt}\selectfont
\caption{Initial circle packing metric}\label{algorithm1}
\textbf{Input:} A triangular mesh $M \in \mathbb{R}^3$.\\
\textbf{Output:} An initial circle packing metric.
\begin{enumerate}
    \item \textbf{for all} $f_{ijk} \in F$:
    \item \quad \quad $ \gamma_i^{jk} = \dfrac{l_{ij}+l_{ki} - l_{jk}}{2} $.
    \item \textbf{end for}
    \item \textbf{for all} $ v_i \in V $:
    \item \quad \quad $ \gamma_i = min_{jk} \gamma_i^{jk}$. 
    \item \textbf{end for}
    \item \textbf{for all} $e_{ij} \in E $:
    \item \quad \quad $ \eta_{ij}=\dfrac{l_{ij}^2-\gamma_i^2-\gamma_j^2}{2\gamma_i\gamma_j} $.
    \item \textbf{end for}
\end{enumerate}
\end{algorithm}

\begin{figure*}[h!]
\centering
\begin{subfigure}{0.4\textwidth}
\includegraphics[width=\textwidth]{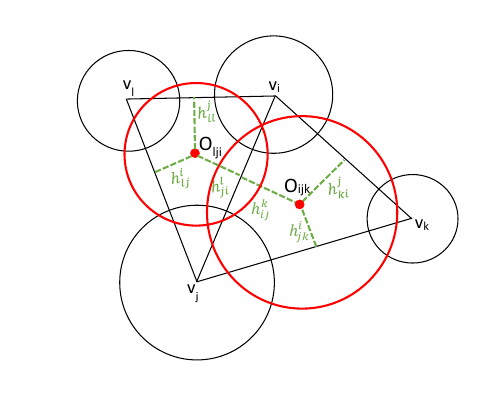}
\caption{}
\label{fig:powerCircle1}
\end{subfigure}
\begin{subfigure}{0.4\textwidth}
\includegraphics[width=\textwidth]{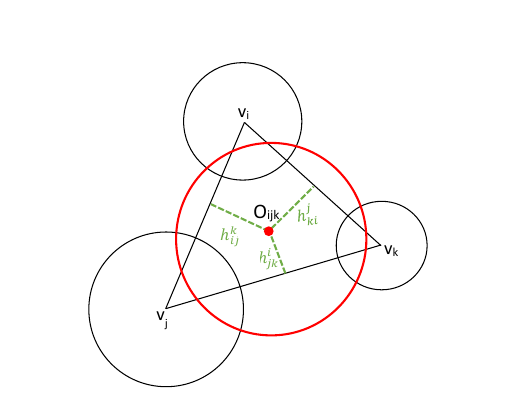}
\caption{}
\label{fig:powerCircle2}
\end{subfigure}
\caption{Power circle. (a): The edge $e_{ij}$ is adjacent to the two faces $f_{ijk}$ and $f_{jil}$, (b): The edge $ e_{ij} $ is attached to a single face $f_{ijk}$.}
\label{fig:powerCircle}
\end{figure*}

\textbf{Ricci energy optimization:}\\
The Ricci energy is a function that measures the deviation of a discrete metric from a desired target, often derived from curvature constraints.\\
After computing the initial circle packing metric, by optimizing the Ricci energy $E(u) = \int{\sum_i(\bar{K_i}-K_i)}du_i $, where $u_i = log \gamma_i$ and $u = (u_1,u_2,...,u_n)^T$ (n is the number of vertices), the metric corresponding to the predefined target curvature is calculated. Traversing the all faces, $f_{ijk} \in E$, we find the power center $o_{ijk} $ which is the center of the circles that are orthogonal to the three vertex circles (see Appendix for computing power center of a triangle). The distance from the power center $o_{ijk} $ to the edges $e_{ij}$ is denoted here by $h_{ij}^k$. By traversing all the edges, if the edge $e_{ij}$ is adjacent to the two faces $f_{ijk}$ and $f_{jil}$, then its weight is given by (see Figure \ref{fig:powerCircle1}): 
\begin{equation} \label{edgeWeightEqu1}
	w_{ij} = \dfrac{h_{ij}^k+h_{ji}^l}{l_{ij}}.
\end{equation}
If the edge $ e_{ij} $ is attached to a single face $f_{ijk}$, then (See Figure \ref{fig:powerCircle2}).
\begin{equation} \label{edgeWeightEqu2}
	w_{ij} = \dfrac{h_{ij}^k}{l_{ij}}.
\end{equation}
The Hessian matrix is then computed as  $H = \dfrac{\partial^2 E}{\partial u_i \partial u_j} $, where 
\begin{equation} \label{HessianEqu}
	\dfrac{\partial^2 E}{\partial u_i \partial u_j} = \begin{cases} 
	\sum_k w_{ik} \quad i = j\\
	-w_{ij} \quad e_{ij} \in M   \quad . \\	
	0 \quad else
	\end{cases}
\end{equation}

In order to optimize the Ricci energy $E(u)$, Newton's method constrained on the hyperplane $\sum_i u_i = 0$ is used, that is:
\begin{equation}
	\delta u = H^{-1}\triangledown E(u) = H^{-1}(\bar{K}-K),
\end{equation} 
where $ \delta u $ is the change in the conformal factor $ u $, $ H^{-1} $ is the inverse of the Hessian matrix, and $  \triangledown E(u)$ is the gradient of the Ricci energy. 
The pipeline of Ricci energy optimization is shown in Algorithm \ref{algorithm2}.

\begin{algorithm}[!h] \fontsize{10pt}{10pt}\selectfont
\caption{Ricci energy optimization}\label{algorithm2}
\textbf{Input:} A triangular mesh $ M $, the target curvature $\bar{K} $.\\
\textbf{Output:} A desired metric $ g $ corresponding to the target curvature $\bar{K} $.
\begin{enumerate}
\item Calculate the initial circle packing metric using Algorithm \ref{algorithm1}.
\item\textbf{while} $ max_{v_i\in M}|\bar{K_i}-K_i|>\epsilon $ \textbf{do}:
\item \quad \textbf{for all} $ f_{ijk} \in F $:
\item \quad \quad Calculate the distance from the power center $ o_{ijk} $ to the edges of $ f_{ijk}$ : $ h_{ij}^k, h_{jk}^i $ and $ h_{ki}^j $.
\item \quad \textbf{end for}
\item \quad \textbf{for all} $e_{ij} \in E $:
\item \quad \quad Calculate the edge weight $ w_{ij} $ using Equations \eqref{edgeWeightEqu1} and \eqref{edgeWeightEqu2}.
\item \quad \textbf{end for}
\item \quad Compute the Hessian matrix using Equation \eqref{HessianEqu}.
\item \quad Solve the linear equation $ H\delta u = \bar{K} - K $ constrained on $ \sum_i u_i = 0  $.
\item   \quad Update the conformal factor $ u \leftarrow u + \delta u $.
\item 	\quad \textbf{for all} $e_{ij} $:
\item \quad \quad Calculate the length of the edge by $ l_{ij}^2 = \gamma_i^2 + \gamma_j^2 + 2\gamma_i\gamma_j\eta_{ij} $.
\item \quad \textbf{end for}
\item   \quad \textbf{for all} $ f_{ijk} \in F$:
\item \quad \quad Calculate the corner angles $ \theta_i^{jk} $, $ \theta_j^{ki} $ and $ \theta_k^{ij} $ based on the cosine law.
\item \quad\textbf{end for}
\item   \quad \textbf{for all} $v_i \in V $:
\item \quad \quad Calculate the Gaussian curvature by Equation \eqref{eqGaussCurv}.
\item \quad \textbf{end for}
\item \textbf{end while}
\end{enumerate}
\end{algorithm}

\textbf{Embedding:}

After obtaining the required metric by Ricci energy optimization, the next step is to embed the surface mesh with normalized metric into a plane $ \mathbb{R}^2 $, to visualize or analyze the geometric properties of the brain mesh.\\
The planar embedding of the mesh is generated via the following pipeline:\\
\textbf{Constructing the suitable metric tensor:} From the Ricci energy optimization configuration, construct the final metric tensor $ g_{ij} $ corresponding to the target curvature $ \bar{K} $.\\
\textbf{Isometric embedding:}
Seek an embedding that preserves distances. First, randomly select an initial face $f_{ijk}$ and flatten it isometrically in the plane:
\begin{equation}\label{embed1}
	\phi(v_0) = 0, \quad \phi(v_1) = l_{01}, \quad \phi(v_2) = l_
 {20}e^{i\theta_0^{12}},
\end{equation}
where, $\phi$ is the isometric embedding function.\\
Then, insert all the neighboring faces in a queue. The head of face $f_{ijk}$ is popped while the queue is not empty. Assume that the isometries $\phi(v_i)$ and $\phi(v_j)$ have been computed. Consequently, $\phi(v_k)$ is at the point where the two circles $(\phi(v_i),l_{ik})$ and $(\phi(v_j),l_{jk})$ cross each other, furthermore, $\phi(v_k)$ is chosen to maintain the direction of triangles. For each face $f_{ijk}$  in the mesh, a queue is created, and all of its neighboring faces that have not yet been flattened, are added to the queue. Until the queue is empty, the procedure of flattening the faces is repeated. Algorithm \ref{algorithm3} outlines the steps for completing this procedure.

\begin{algorithm}[!h] \fontsize{10pt}{10pt}\selectfont
\caption{Embedding into the plane}\label{algorithm3}
\textbf{Input:} A triangular mesh $ M $.\\
\textbf{Output:} An isometric embedding $ \phi $.
\begin{enumerate}
\item Embed a randomly selected face $ f_{ijk} \in F $ using Equation \eqref{embed1}.
\item Create the queue $Q$ with all of the initial face's neighboring faces.
\item \textbf{while} $\mid Q \mid > 0$ \textbf{do}:
\item \quad Pop the head face $ f_{ijk} $ from the queue $Q$.
\item \quad Assume that $ v_i $ and $ v_j $ were embedded in the plane, compute the points where the two circles intersect, $ (\phi(v_i),l_{ik}) \cap (\phi(v_j),l_{jk}) $. 
\item \quad Add to the queue the neighborhood faces of $ f_{ijk}$ that have not yet been accessed.
\item \textbf{end while}
\end{enumerate}
\end{algorithm}

\subsubsection{Feature extraction}
Geometric features were derived from the mesh through a two-stage process: first, Gaussian curvature was calculated directly on the surface mesh; second, we used Euclidean Ricci flow to calculate a planar conformal parametrization, establishing a conformal mapping between a surface mesh with a boundary and a flat plane. Following this, we assessed the Area Distortion and Conformal Factor features for every vertex. 

\textbf{Gaussian curvature:} Gaussian curvature is an intrinsic geometric property that quantifies the local bending of a surface at each point, defined as the product of the principal curvatures. In 3D image classification, it serves as a discriminative feature by capturing surface complexity, distinguishing between convex, concave, and saddle-shaped regions. Unlike mean curvature, which only measures average bending, Gaussian curvature provides invariance to local isometric deformations, making it robust to certain shape variations. By integrating Gaussian curvature into feature descriptors, classifiers can better differentiate between objects with similar global shapes but distinct local geometries, such as biological structures or manufactured parts. Its computation on discrete meshes or point clouds, often via normal variation or quadratic surface fitting, enables effective shape analysis in applications like medical imaging and object recognition.

\begin{figure}
    \centering
    \begin{subfigure}{0.48\textwidth}
    \includegraphics[width=\textwidth]{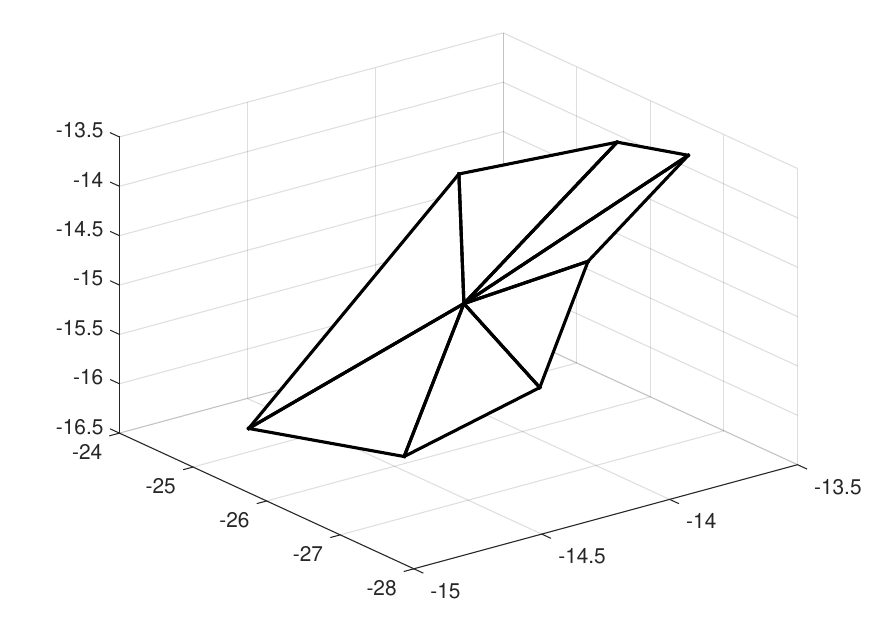}
     \caption{}
    \end{subfigure}
    \begin{subfigure}{0.45\textwidth}
    \includegraphics[width=\textwidth]{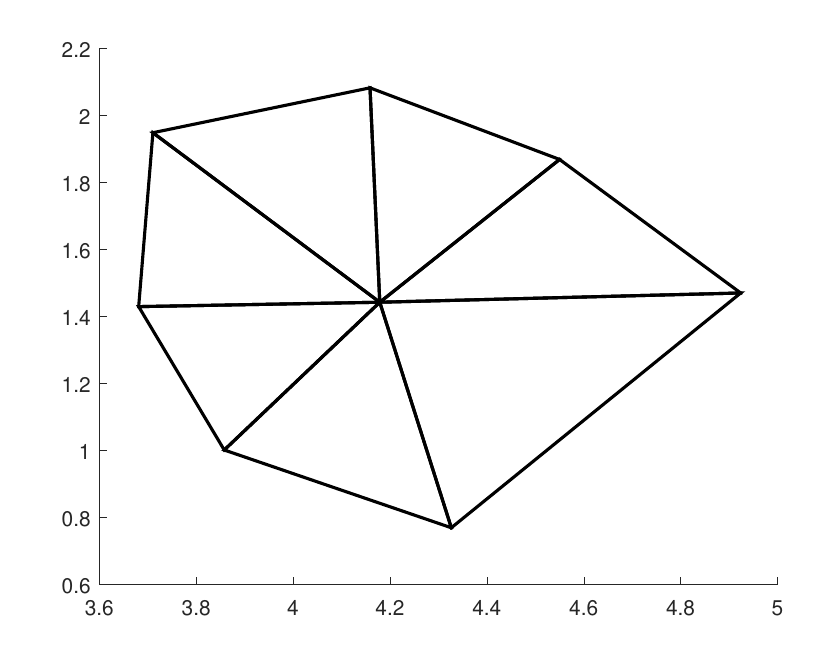}
     \caption{}
    \end{subfigure}
    \caption{(a): The region surrounding a vertex in the initial stage (3-dimensional), (b): The region surrounding a vertex in the final stage (2-dimensional) of Ricci energy optimization.}
    \label{localArea}
\end{figure}

\textbf{Area distortion:} As previously mentioned, the Ricci flow alters the Riemannian metric based on curvature, evolving like a heat diffusion process until it becomes uniform across the surface. Consequently, varying curvatures at the vertices result in different metrics along the edges. We can utilize this metric as a feature by calculating the area of triangles within the one-ring local neighborhood of each vertex on the mesh during both the initial and current stages of optimization. We then determine the difference in local areas between these two stages, which we refer to as the \textit{Area Distortion}, as shown in equation \ref{areaDistortion}.
\begin{equation} \label{areaDistortion}
AD(v) = \sum_{t \in B} area(t)-\sum_{t \in B} \widehat{area}(t),
\end{equation}
where $ area(t) $ represents the area of triangle $ t $ on the initial mesh, $ \widehat{area} $ denotes the area of triangle $ t $ on the mesh in the current stage, and $ B $ refers to the one-ring neighborhood of vertex $ v $. Figure \ref{localArea} shows the local one-ring neighborhood of a vertex in both the first and last stages of Ricci flow optimization. 

\textbf{Conformal factor:} The Ricci flow method attains a conformal parameterization by transforming a 2-manifold into a Euclidean plane, a 2-sphere, or a hyperbolic plane, based on the topology of the manifold, resulting in constant Gaussian curvature (0, +1, or -1, respectively). Initially, circle packing metrics define the Gaussian curvature, which is then aligned with the target curvature through Ricci flow. The process modifies the radii of the circles, changing the Gaussian curvature until it becomes uniform across all vertices.
In this process, the conformal factor at the vertex $v_i$, is defined as $u_i=\log(\gamma_i)$ where $\gamma_i$ represents the radius of the circle at that vertex. By definition, $u_i$ captures intrinsic surface information and remains unchanged under isometries. 
\subsection{Computing the entropy of features}\label{sec4}
Shannon entropy serves as a valuable tool, not only in understanding the uncertainty of information but also in analyzing geometric structures such as triangulation meshes. In computational geometry, the complexity of a mesh can significantly impact rendering and processing efficiency. By applying Shannon entropy to triangulation meshes, we can assess the level of detail and irregularity in the mesh's structure. In the following, a step-by-step method for using Shannon entropy to index the irregularity of a triangulation mesh is presented:\\
\textbf{Step 1: Compute the range of features:} The possible values that the feature can be established by. \\
\textbf{Step 2: Partition the attribute range:} The range of the feature is divided into bins or intervals, with the number of bins adjustable according to the desired level of granularity and sensitivity to irregularities. This approach ensures that important areas are not overlooked, allowing for the capture of both small, distinctive details and large, interesting regions. The optimal intervals are selected through a trial-and-error process.\\
\textbf{Step 3: Compute the probability distribution of the feature:} The probability distribution of the attribute values is computed by determining the frequency or proportion of vertices that fall into each bin. This is achieved by counting the number of vertices in each bin and dividing that count by the total number of vertices. The mathematical representation of this process can be expressed as: 
\begin{equation} \label{eqProb}
    p_i=\frac{n_i}{n},
\end{equation}
where $ p_i $ is the probability of the feature value in bin $ i $, $ n_i $ is the number of vertices in bin $ i $, and $ n $ is the total number of vertices across all bins.\\
\textbf{Step 4: Compute Shannon entropy:} Once you have the probability distribution $ p_i $, you can compute the Shannon entropy using the following equation:
 \ref{eqEntropy}:
\begin{equation} \label{eqEntropy}
    E = -\sum_{i=1}^n p_i \log p_i,
\end{equation}
where $ E $ represents the Shannon entropy, $ p_i $ is the probability of the attribute value in bin $ i $, and the logarithm is typically taken in base 2 (though natural logarithm can also be used, depending on the context). This measure quantifies the uncertainty or information content associated with the distribution of the attribute values.

By applying Shannon entropy to the triangulation mesh, we can derive a numerical indicator of irregularity, which can be utilized for indexing or comparing various meshes. In this paper, we focus on surface indexing through Ricci flow, incorporating Shannon entropy into our analysis. This approach allows us to convert and compress the information obtained from Ricci flow parameterization, enabling us to represent it more concisely. 

\begin{figure}[!h]
    \centering   
    \includegraphics[width= 0.9\textwidth]{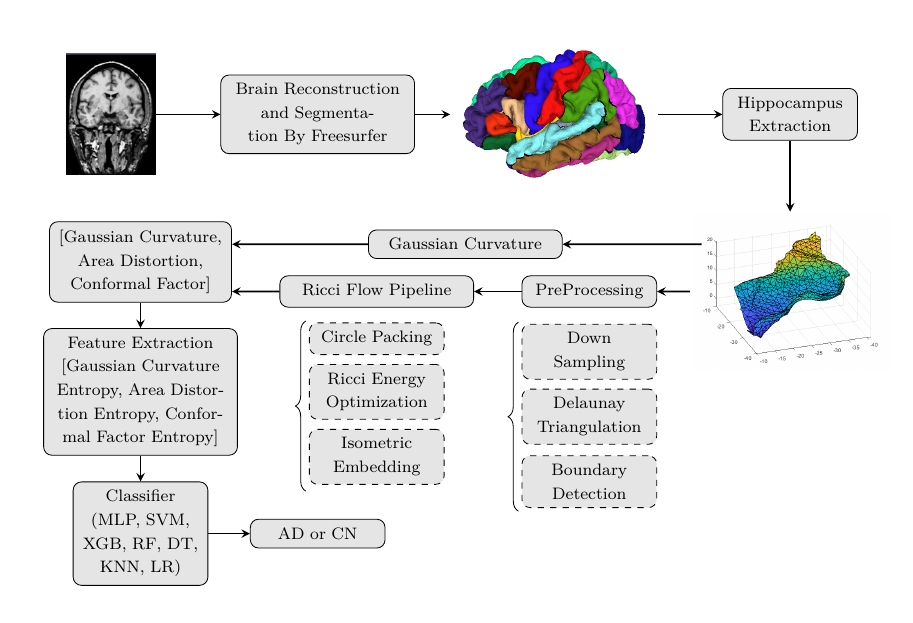}  \caption{Block diagram of the proposed method in this paper.}
    \label{blockDiagram}
\end{figure}

Figure \ref{blockDiagram} displays a block representation of the suggested technique, illustrating the various components and their interactions. The algorithm outlining this process is detailed in Algorithm \ref{algorithm4}, which provides a step-by-step framework for implementing the technique effectively. The combination of the block diagram and the algorithm serves to clarify the methodology and facilitate understanding of the approach proposed in this work.

\begin{algorithm}[!h] \fontsize{10pt}{10pt}\selectfont
\caption{Alzheimer's disease diagnosis using Ricci flow and classification models}\label{algorithm4}
\textbf{Input:} An MRI data.\\
\textbf{Output:} The class of input MRI data.
\begin{enumerate}
\item Use the Freesurfer automated processing pipeline to reconstruct the brain surface triangulation mesh from MRI data.
\item Extract the hippocampal regions of both hemispheres from the brain surface triangulation mesh.
\item Equalize the number of vertices in the triangulation mesh of hippocampal regions by down-sampling (or up-sampling).
\item Change the triangulation mesh of the hippocampal regions to the Delaunay triangulation mesh.
\item Find the boundary of the meshes of the hippocampal regions.
\item Calculate the inversive distance circle packing metric of each mesh using Algorithm \ref{algorithm1}.
\item Optimize the Ricci energy of each mesh by Newton's method using Algorithm \ref{algorithm2}.
\item Embed each mesh with a normalization metric in the plane using Algorithm \ref{algorithm3}.
\item Calculate the area distortion, Gaussian curvature, and conformal factor attributes of each vertex $v$ on the triangulation meshes.
\item Find a qualified uniform sampling of attributes computed into an appropriate number of bins for each triangulation mesh.
\item Calculate the entropy of attributes computed for each triangulation mesh of the hippocampal regions by Equation \eqref{eqEntropy} as a feature vector for input data.
\item Classify the input data using the classifier (XGBoost, Random Forest, SVM, MLP, Decision Tree, KNN, Logistic Regression) based on the feature vector achieved.
\end{enumerate}
\end{algorithm}
\subsection{Classification}
In the classification phase, we used various classifiers, including: 

\textbf{Support Vector Machine (SVM):} Support Vector Machine is s supervised learning model used for classification and regression tasks. SVM works by finding the hyperplane that best separates the classes in a high-dimensional space. It aims to maximize the margin between different classes, effectively creating boundaries that can classify new data points. SVM is effective in high-dimensional spaces and is particularly useful when there is a clear margin of separation between classes.

\textbf{Extreme Gradient Boosting (XGBoost):} XGBoost is an optimized implementation of the gradient boosting framework designed for speed and performance. It builds models sequentially by adding new trees that correct the errors made by previous ones. XGBoost includes regularization techniques to prevent overfitting and can handle missing values natively, making it highly effective for structured data problems as well as image classification when combined with feature engineering \citep{chehreh2025modeling}.

\textbf{Multi-Layer Perceptron (MLP):} Multi-Layer Perceptron is a type of artificial neural network composed of multiple layers: an input layer, one or more hidden layers, and an output layer. MLPs use backpropagation for training, optimizing weights through gradient descent. They are capable of capturing non-linear relationships due to their architecture with activation functions like ReLU or sigmoid. MLPs have been widely adopted in image classification tasks due to their ability to learn hierarchical representations.

\textbf{Decision Tree:} Decision Trees are intuitive tree-like structures used for both classification and regression tasks. They make decisions based on asking simple questions about feature values at each node until they reach a leaf node where predictions are made. The tree is built by recursively splitting the dataset based on feature values that yield the maximum information gain or minimize impurity (like Gini impurity or entropy). Decision Trees offer straightforward interpretation and visualization; however, they may be susceptible to overfitting, particularly when the trees are deep.

\textbf{Random Forest:} Random Forest is an ensemble learning method based on decision trees. It constructs multiple decision trees during training and outputs the mode of their predictions (for classification) or the mean prediction (for regression). By combining multiple trees, Random Forest reduces overfitting and increases robustness, making it effective for various datasets, including those with noisy features or complex relationships \citep{nasiri2024machine}.

\textbf{K-Nearest Neighbors (KNN): }K-Nearest Neighbors is a simple, instance-based learning algorithm used for classification and regression. It classifies a data point based on how its neighbors are classified; specifically, it assigns the class most common among its k nearest neighbors in the feature space. KNN is non-parametric, meaning it makes no assumptions about data distribution. However, it can be computationally expensive as it requires distance calculations for all training samples during prediction.

\textbf{Logistic Regression:} Logistic Regression is a statistical model commonly used for binary classification problems. It models the probability of an input belonging to a particular category using a logistic function (sigmoid function), which outputs values between 0 and 1. The model estimates coefficients through maximum likelihood estimation that reflect the relationship between input features and their log-odds of class membership. Despite its name, Logistic Regression is primarily used for classification tasks rather than regression.

\section{Results}
The efficacy and effectiveness of our method were demonstrated through the examination of 3D MRI data from individuals with Alzheimer’s disease (AD) as well as those recognized as cognitively normal (CN).
Our method’s efficiency and effectiveness was showcased through the analysis of human brain cortices from individuals with Alzheimer’s disease (AD) and those who are cognitively normal (CN). The computational tasks were carried out using MATLAB and Python on a laptop running Windows 11, powered by a  2.10 GHz 13th Gen Intel(R) Core(TM) i5-13420H and 16 GB of RAM.

\begin{figure*}[h!]
\centering
\begin{subfigure}{0.49\textwidth}
\includegraphics[width=\textwidth]{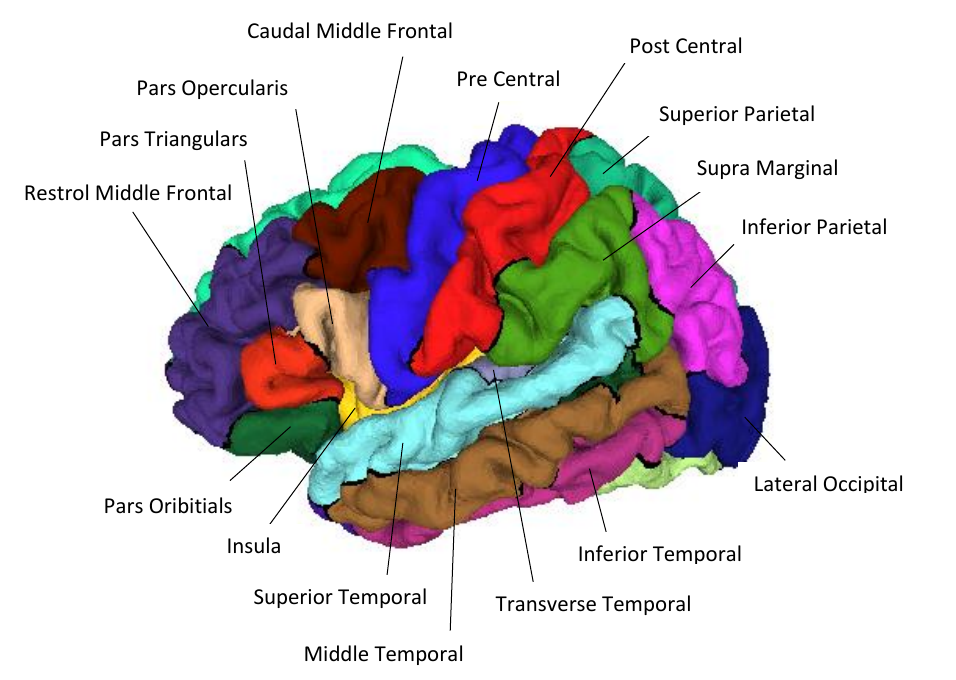}
\caption{superior view}
\end{subfigure}
\begin{subfigure}{0.49\textwidth}
\includegraphics[width=\textwidth]{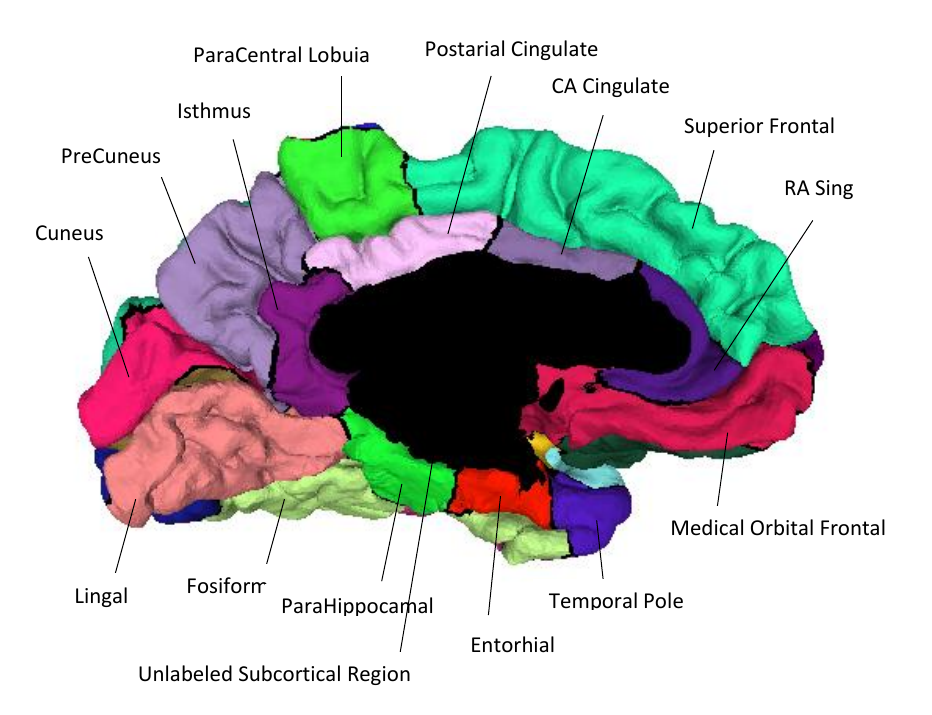}
\caption{Inferior view}
\end{subfigure}
\caption{Depiction of the functional regions of the left hemisphere of the cerebral cortex.}
\label{brainColoring}
\end{figure*}

The data utilized in the creation of this paper was sourced from the Alzheimer’s Disease Neuroimaging Initiative (ADNI) database\footnote{https://adni.loni.usc.edu/data-samples/access-data/} \citep{jack2008alzheimer}.\\ 
We employed the Freesurfer automated processing pipeline, as introduced in \citep{dale1999cortical}, for tasks such as automatic skull stripping, tissue categorization, extraction of surfaces, and parcellation of cortical and subcortical regions. This pipeline calculates geometric attributes like curvature, curving, and local folding for each parcellation and provides data on surface and volume for approximately 34 distinct cortical structures \citep{desikan2006automated}. Figure \ref{brainColoring} presents the left hemisphere of the brain, showcasing a number of functional areas.

\begin{figure}[!h]
    \centering
    \begin{subfigure}{0.48\textwidth}
    \includegraphics[width=\textwidth]{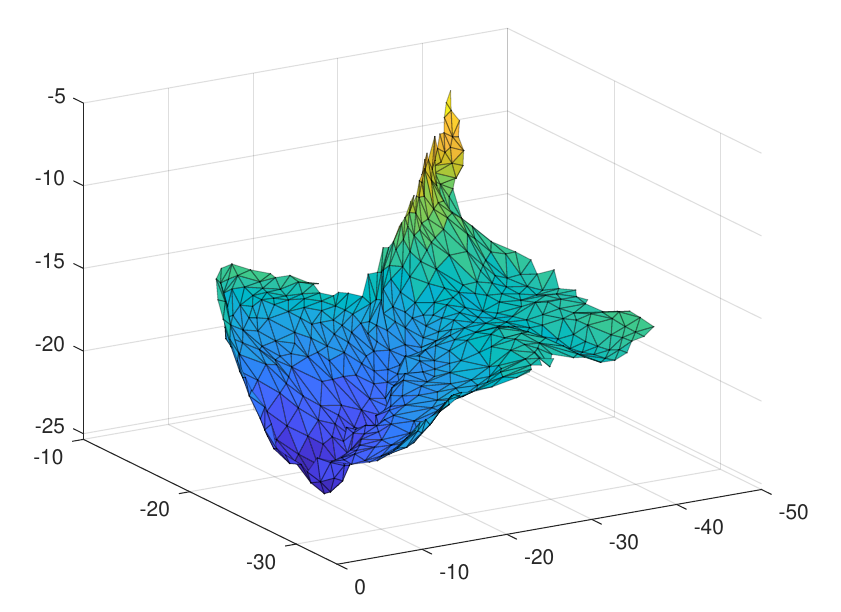}
    \caption{}    
    \end{subfigure}
    \begin{subfigure}{0.48\textwidth}
    \includegraphics[width=\textwidth]{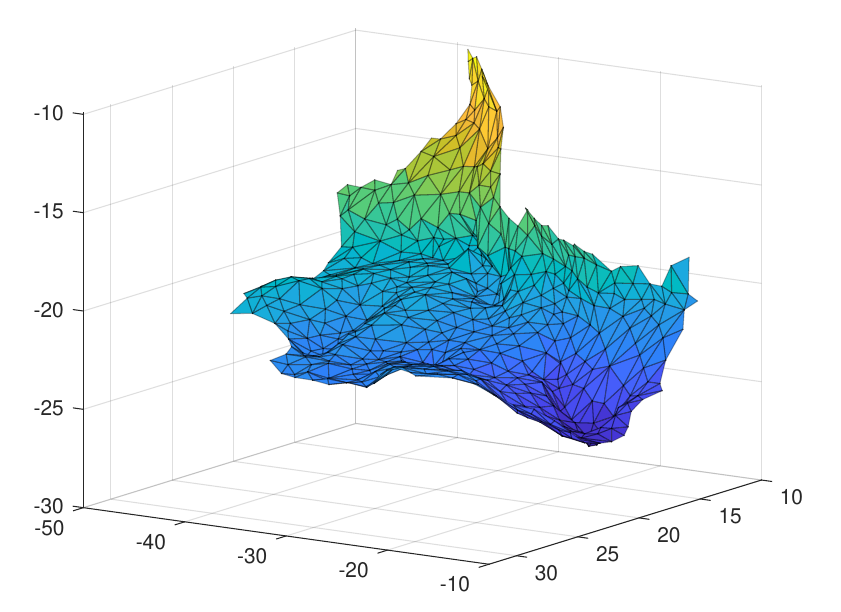}
    \caption{}    
    \end{subfigure} 
    \caption{The hippocampal regions located on the left (a) and right (b) sides of a subject's cortical hemisphere.}
    \label{fig:hippocampal regions of a AD subject}    
\end{figure}

\begin{figure}[!h]
    \centering
    \begin{subfigure}{0.48\textwidth}
    \includegraphics[width=\textwidth]{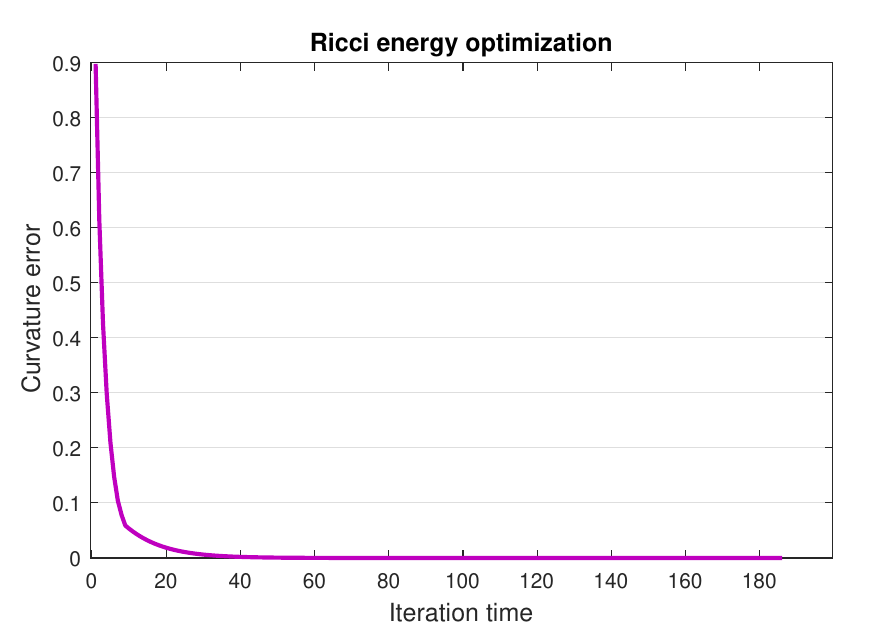}
    \caption{}     
    \end{subfigure}
    \begin{subfigure}{0.48\textwidth}
    \includegraphics[width=\textwidth]{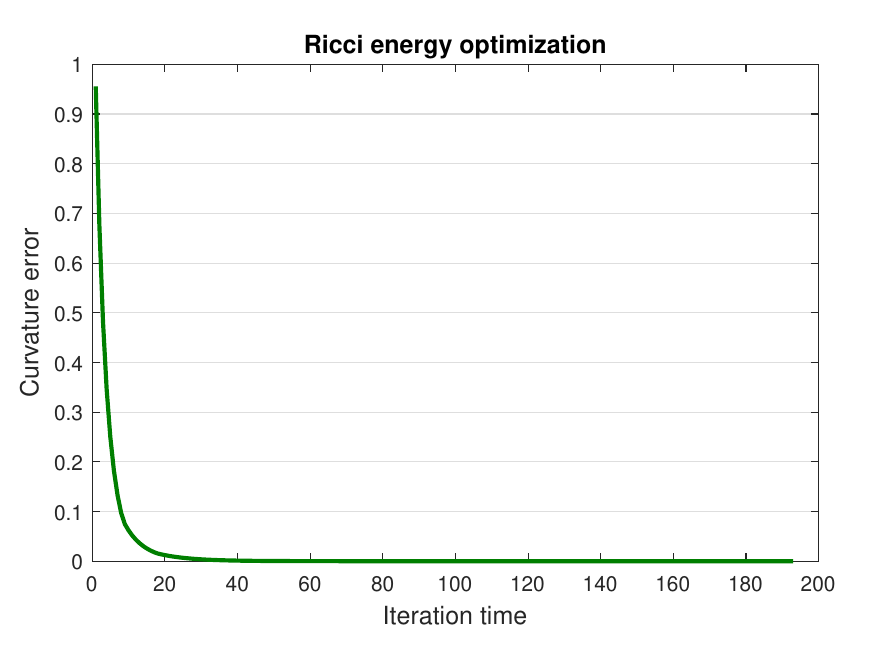}
    \caption{}    
    \end{subfigure}    
    \caption{Optimization of Ricci energy across a subject's left(a) and right(b) hippocampal regions.}
    \label{fig:convergence of a AD subject}
\end{figure}

\begin{figure}
    \begin{subfigure}{0.46\textwidth}
    \includegraphics[width=\textwidth]{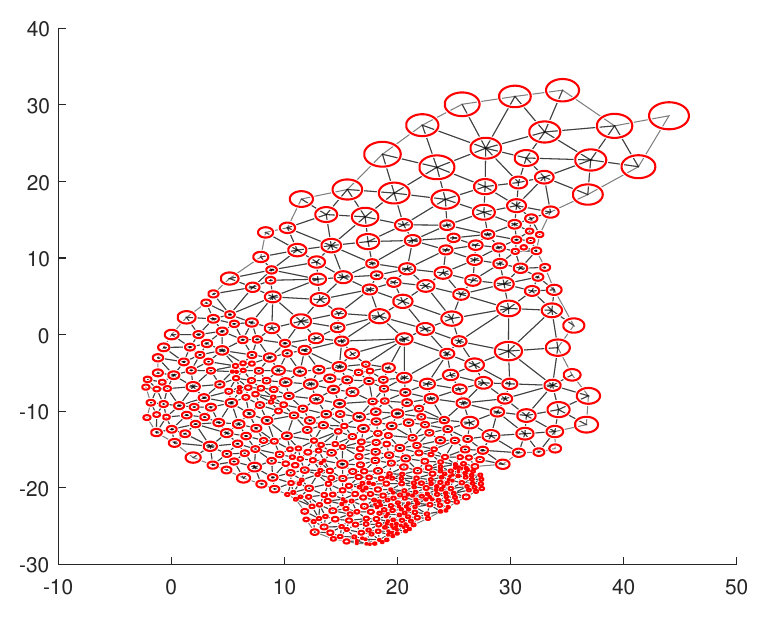}
    \caption{}    
    \end{subfigure}
    \begin{subfigure}{0.46\textwidth}
    \includegraphics[width=\textwidth]{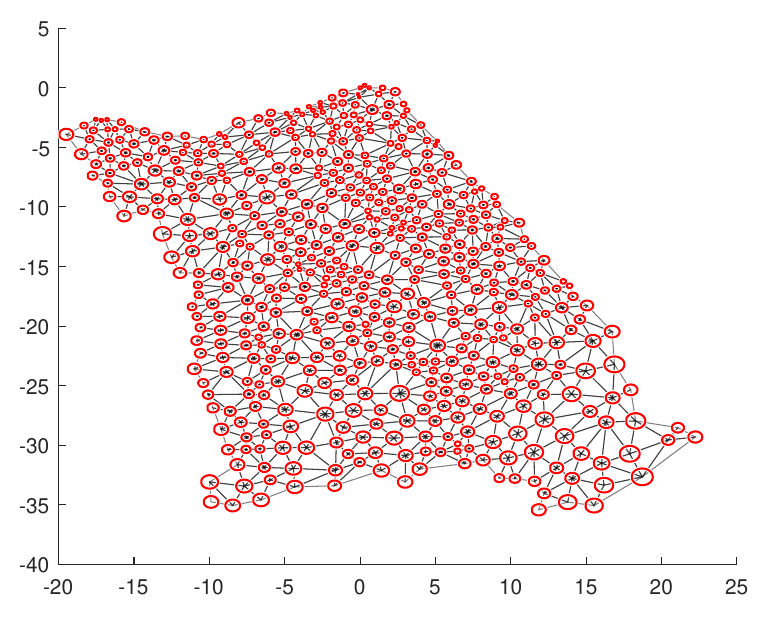}
    \caption{}    
    \end{subfigure}
    \caption{Planar embedding of the left(a) and right(b) hippocampal regions of a subject. The red circles in (a) and (b) represent the inversive distance circle packing in the normalization space.}
    \label{fig:embedding of a AD subject}
\end{figure}

\begin{figure*}[!h]
\centering
\begin{subfigure}{0.31\textwidth}
\includegraphics[width=\textwidth]{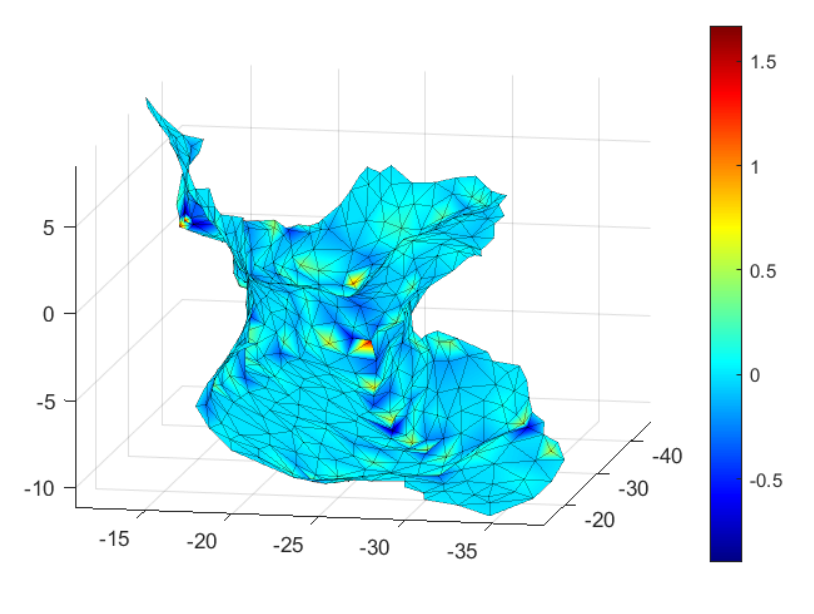}
\caption{}
\end{subfigure}
\begin{subfigure}{0.27\textwidth}
\includegraphics[width=\textwidth]{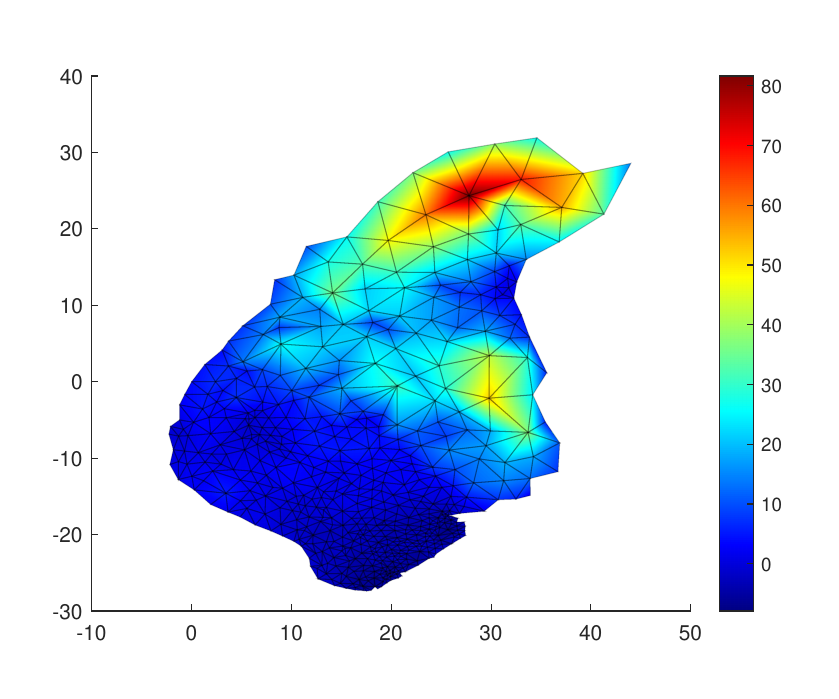}
\caption{}
\end{subfigure}
\begin{subfigure}{0.27\textwidth}
\includegraphics[width=\textwidth]{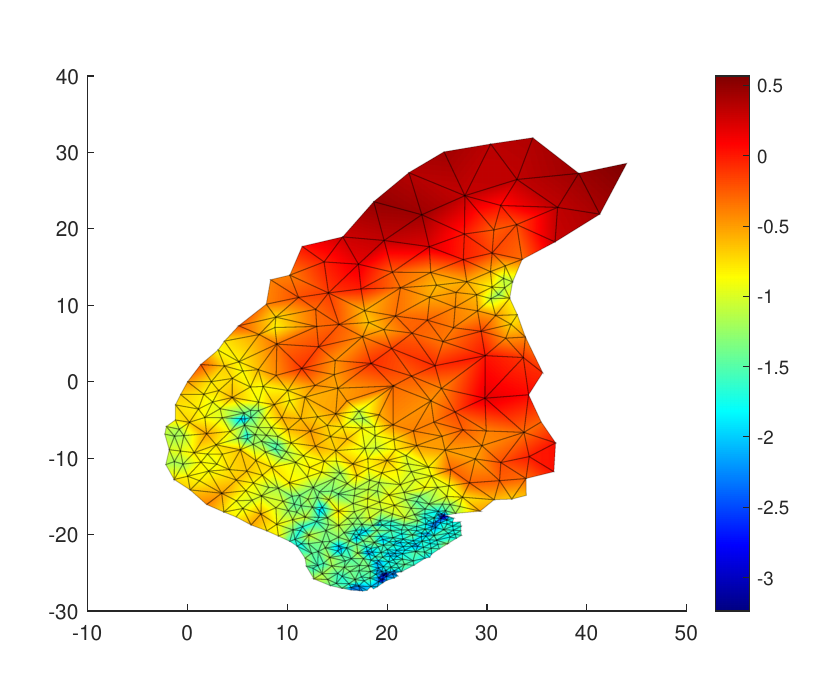}
\caption{}
\end{subfigure}
\begin{subfigure}{0.31\textwidth}
\includegraphics[width=\textwidth]{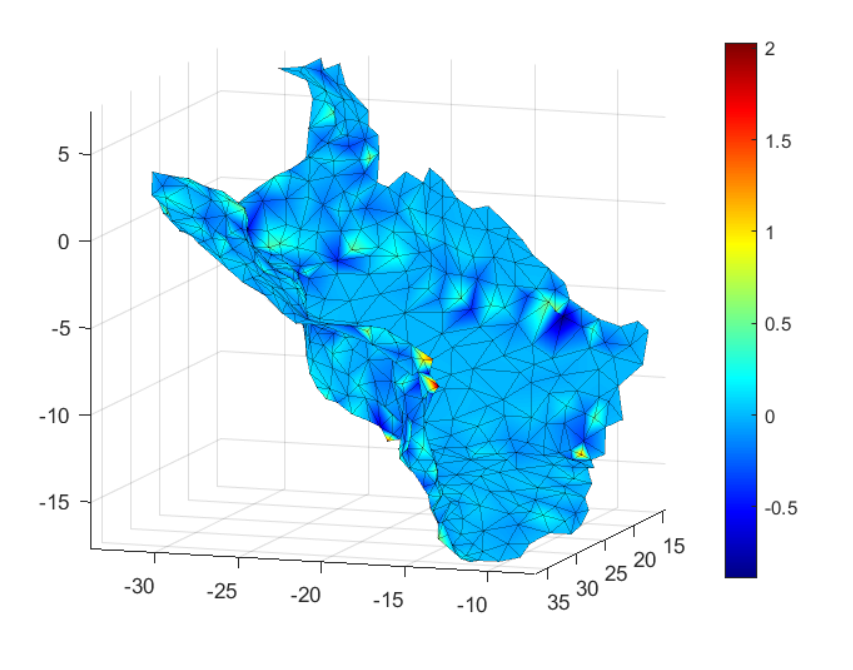}
\caption{}
\end{subfigure}
\begin{subfigure}{0.27\textwidth}
\includegraphics[width=\textwidth]{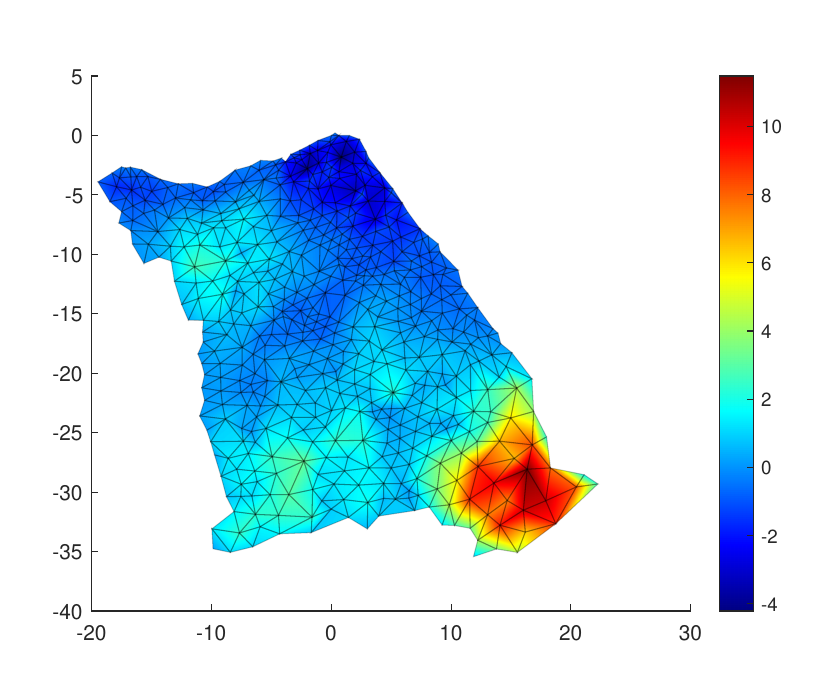}
\caption{}
\end{subfigure}
\begin{subfigure}{0.27\textwidth}
\includegraphics[width=\textwidth]{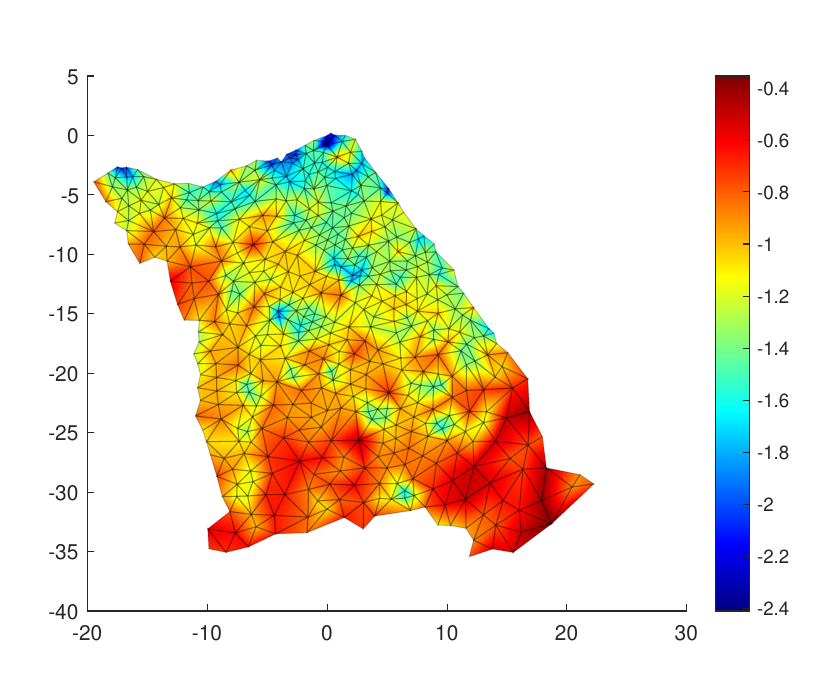}
\caption{}
\end{subfigure}
\caption{Distribution of features area distortion, Gaussian curvature, and conformal factor on the left (a,b,c) and right (d,e,f) of the hippocampus region of an AD subject.}
\label{fig:disFeaturesAD}
\end{figure*}

\begin{figure*}[!h]
\centering
\begin{subfigure}{0.31\textwidth}
\includegraphics[width=\textwidth]{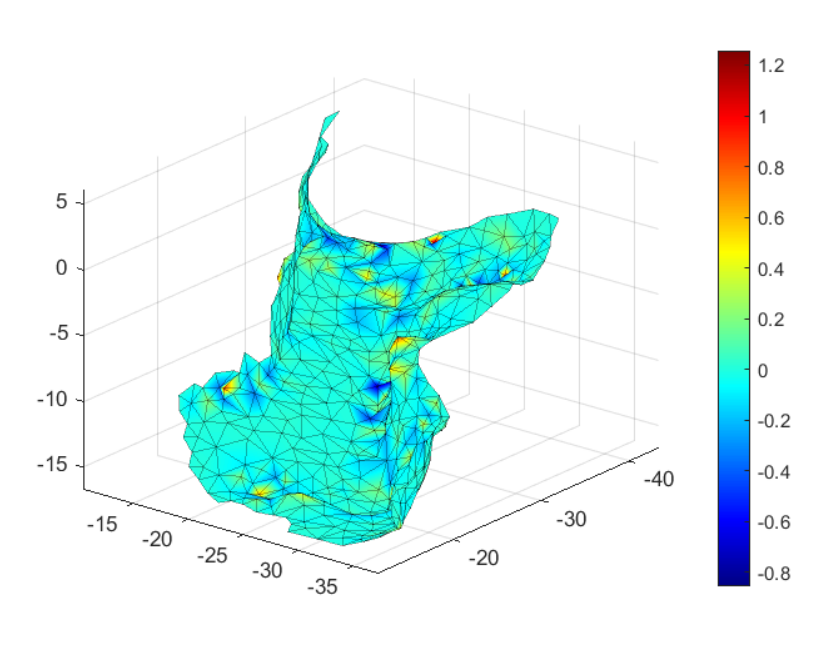}
\caption{}
\end{subfigure}
\begin{subfigure}{0.27\textwidth}
\includegraphics[width=\textwidth]{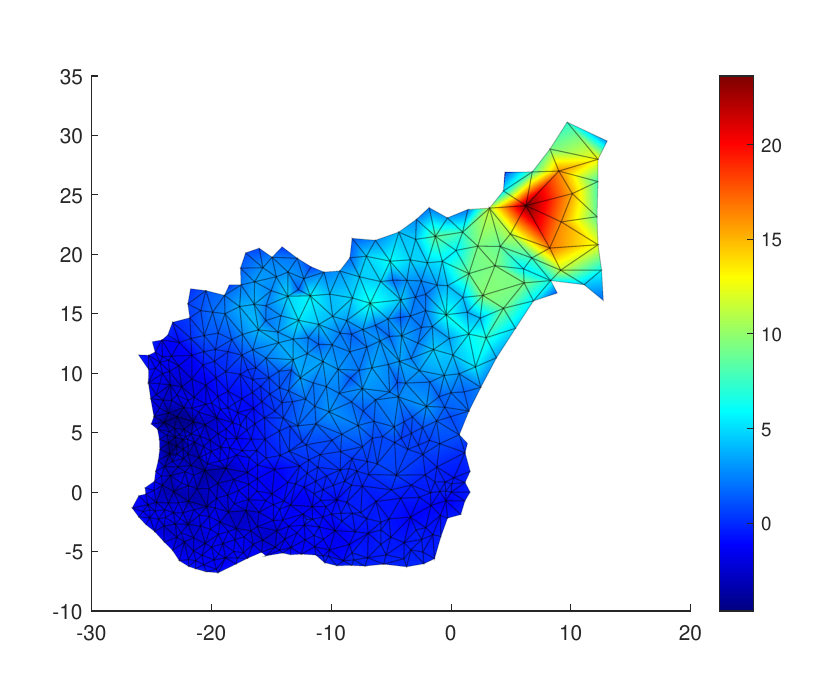}
\caption{}
\end{subfigure}
\begin{subfigure}{0.29\textwidth}
\includegraphics[width=\textwidth]{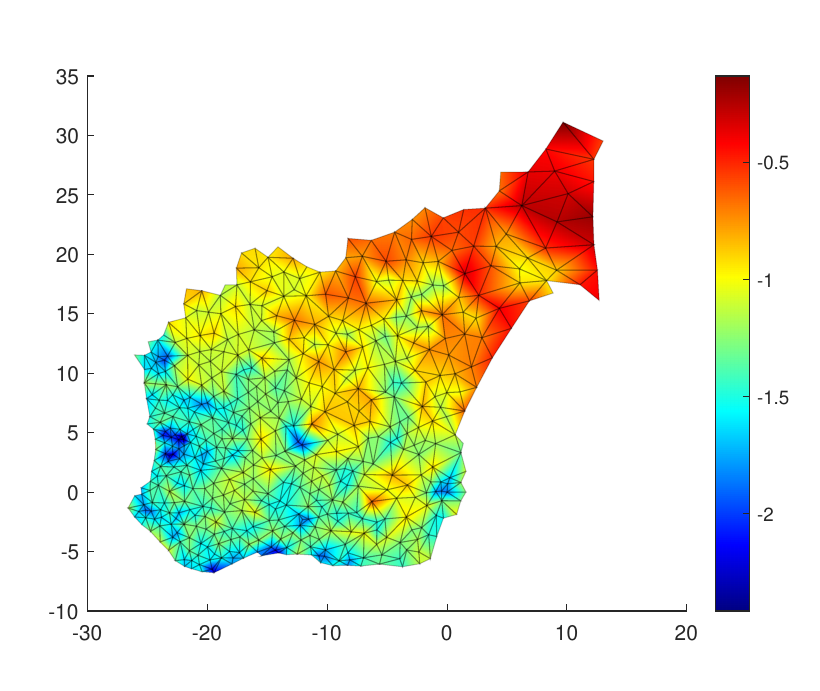}
\caption{}
\end{subfigure}
\begin{subfigure}{0.29\textwidth}
\includegraphics[width=\textwidth]{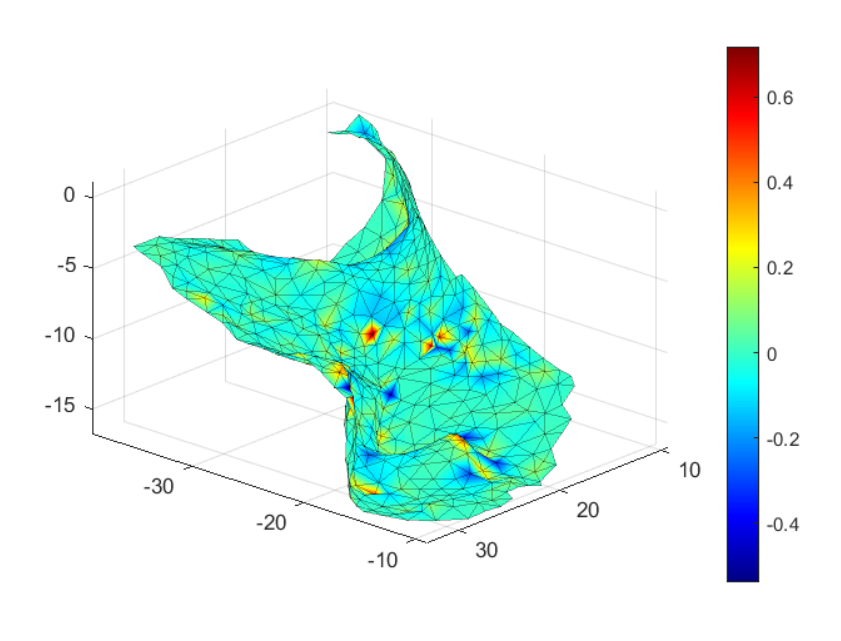}
\caption{}
\end{subfigure}
\begin{subfigure}{0.29\textwidth}
\includegraphics[width=\textwidth]{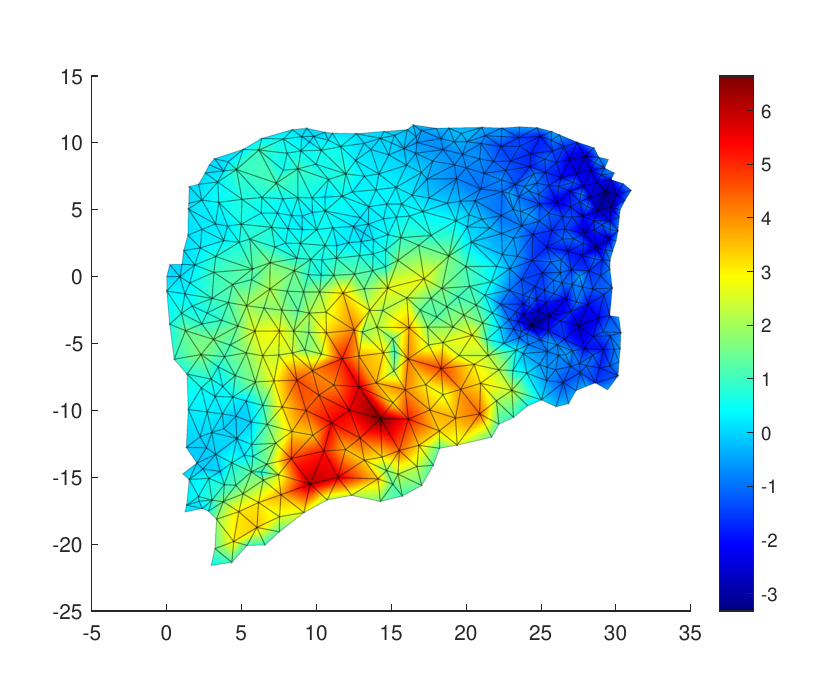}
\caption{}
\end{subfigure}
\begin{subfigure}{0.29\textwidth}
\includegraphics[width=\textwidth]{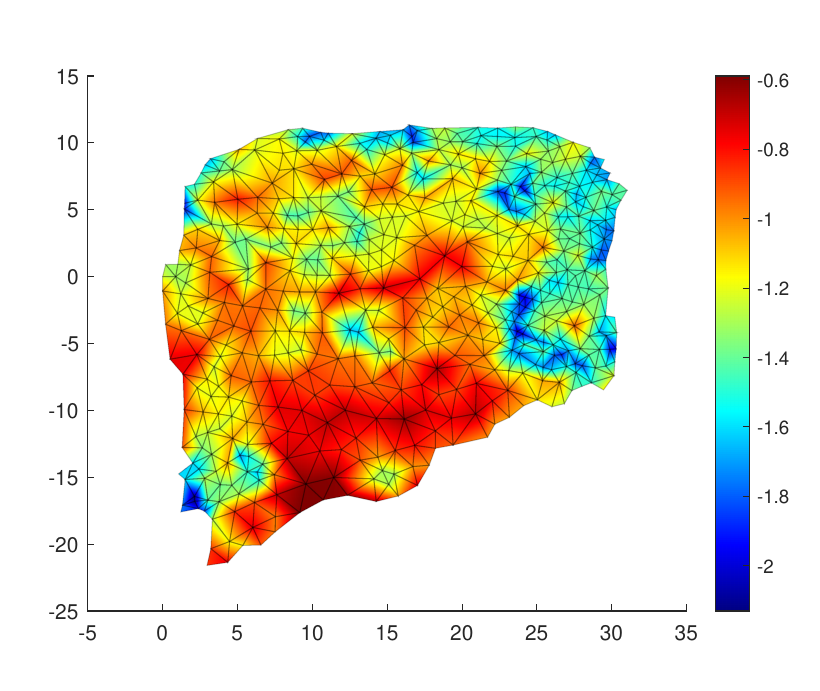}
\caption{}
\end{subfigure}
\caption{Distribution of features area distortion, Gaussian curvature, and conformal factor on left (a,b,c) and right (d,e,f) of hippocampus region of a CN subject.}
\label{fig:disFeaturesCN}
\end{figure*}

The hippocampal areas from both hemispheres of the brain's surface were originally obtained through parcellation. The hippocampal areas within an individual's left and right hemispheres are visible in Figure \ref{fig:hippocampal regions of a AD subject}.
Subsequently, each region was subjected to the Ricci flow method. The Ricci energy optimization on each region and its planar embeddings are illustrated in Figures \ref{fig:convergence of a AD subject} and \ref{fig:embedding of a AD subject}, respectively.
Subsequently, we computed the signatures of area distortion, Gaussian curvature, and conformal factor at the vertices located within the canonical domain space. The characteristics of area distortion, Gaussian curvature, and conformal factor distribution across the triangulation mesh in the canonical domain space of the left and right hippocampal regions for both an Alzheimer's disease (AD) subject and a cognitively normal (CN) subject are depicted in Figures \ref{fig:disFeaturesAD} and \ref{fig:disFeaturesCN}, respectively. It is evident that each of these features has a distinct distribution within the hippocampal region. 
\newpage
We employed Shannon entropy to quantify this significant difference by calculating the entropy associated with area distortion, the conformal factor, and Gaussian curvature. These calculations have been designated as \textit{area distortion entropy}, \textit{conformal factor entropy}, and \textit{Gaussian curvature entropy}, respectively, serving as the proposed signatures. 

\begin{figure}
    \begin{subfigure}{0.3\textwidth}
    \includegraphics[width=\textwidth]{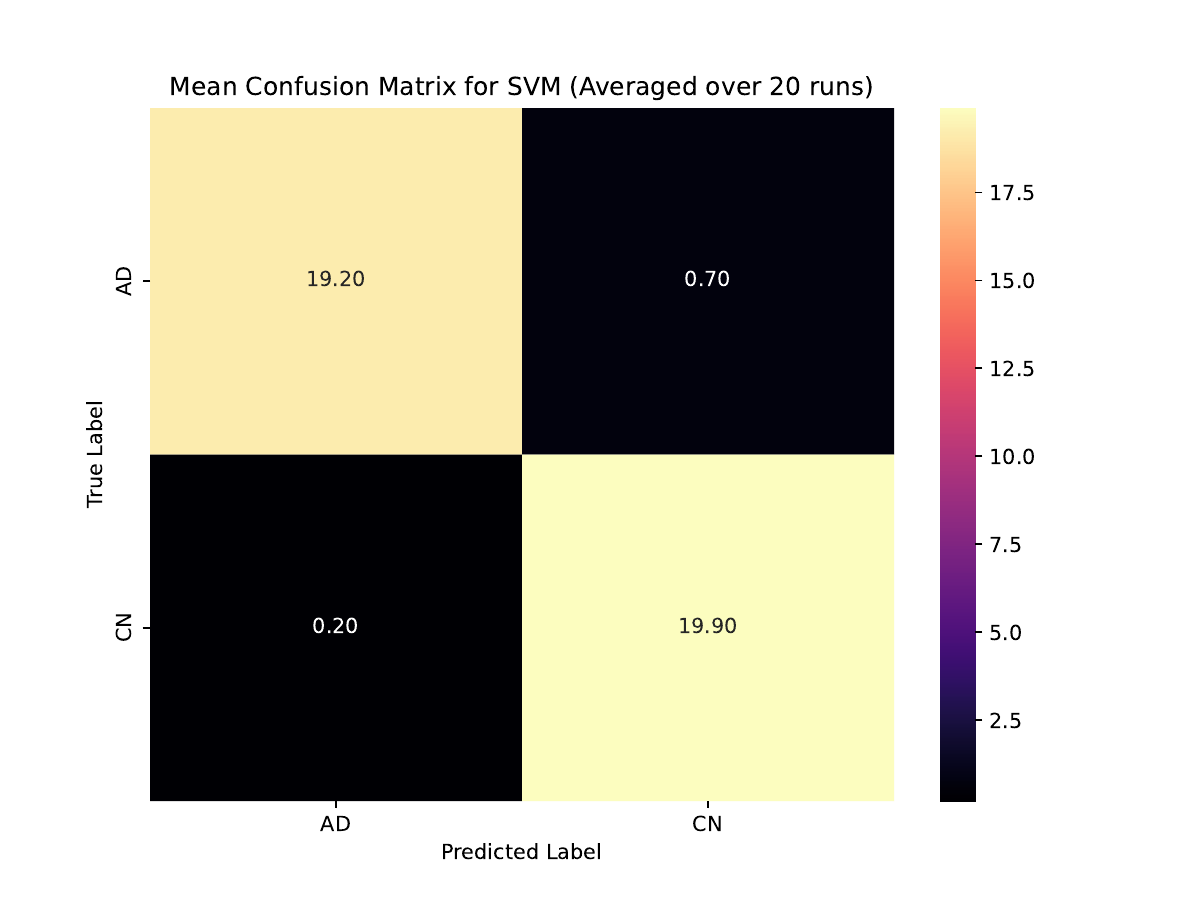}
    \caption{SVM}    
    \end{subfigure}
    \begin{subfigure}{0.3\textwidth}
    \includegraphics[width=\textwidth]{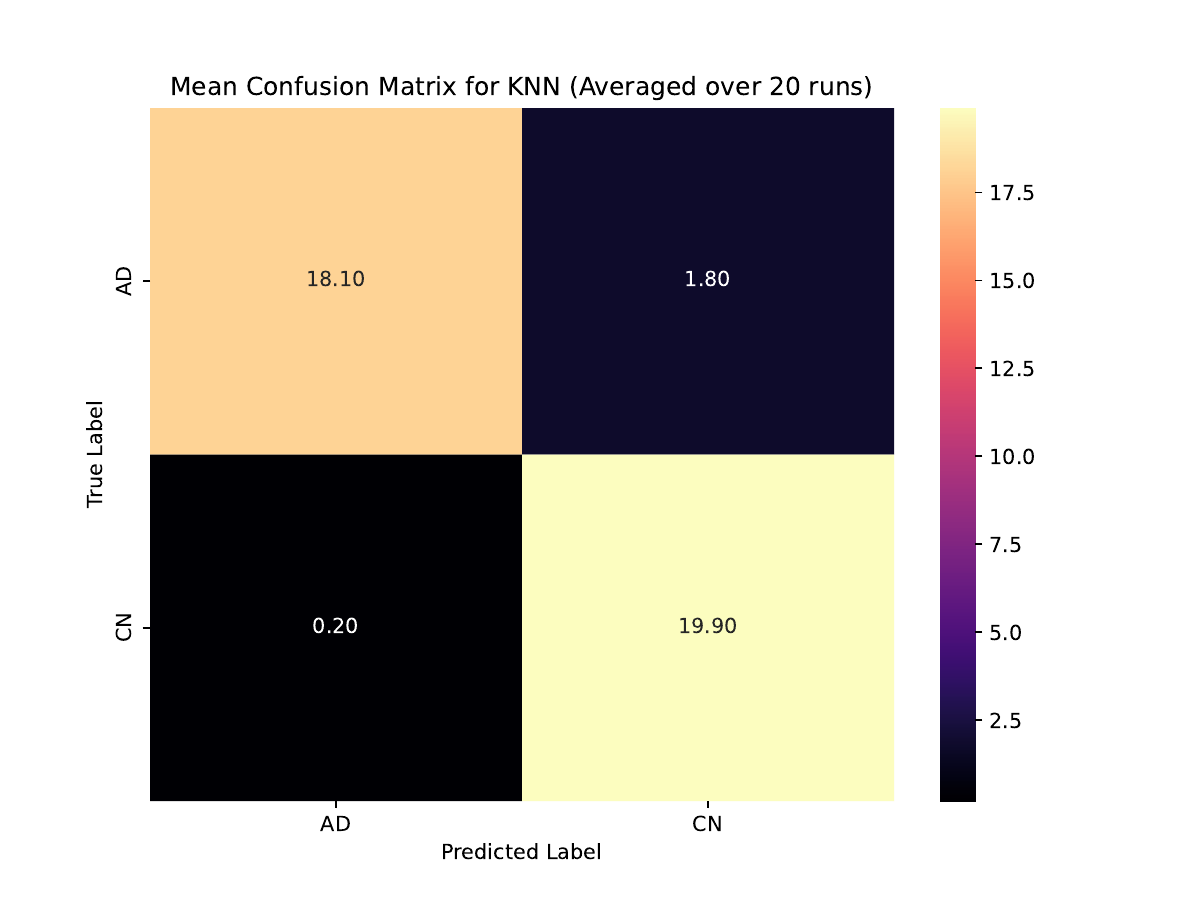}
    \caption{KNN}    
    \end{subfigure}
    \begin{subfigure}{0.3\textwidth}
    \includegraphics[width=\textwidth]{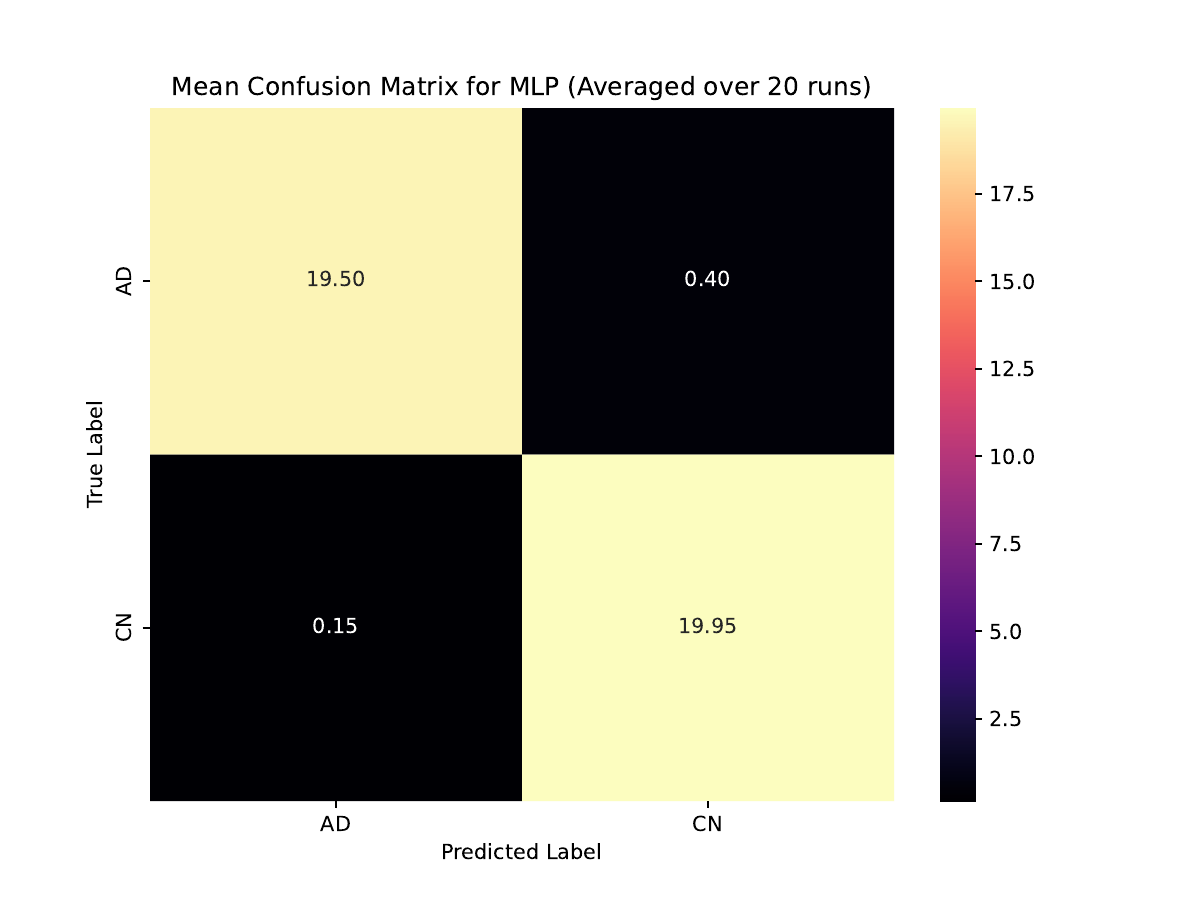}
    \caption{MLP}    
    \end{subfigure}
    \begin{subfigure}{0.3\textwidth}
    \includegraphics[width=\textwidth]{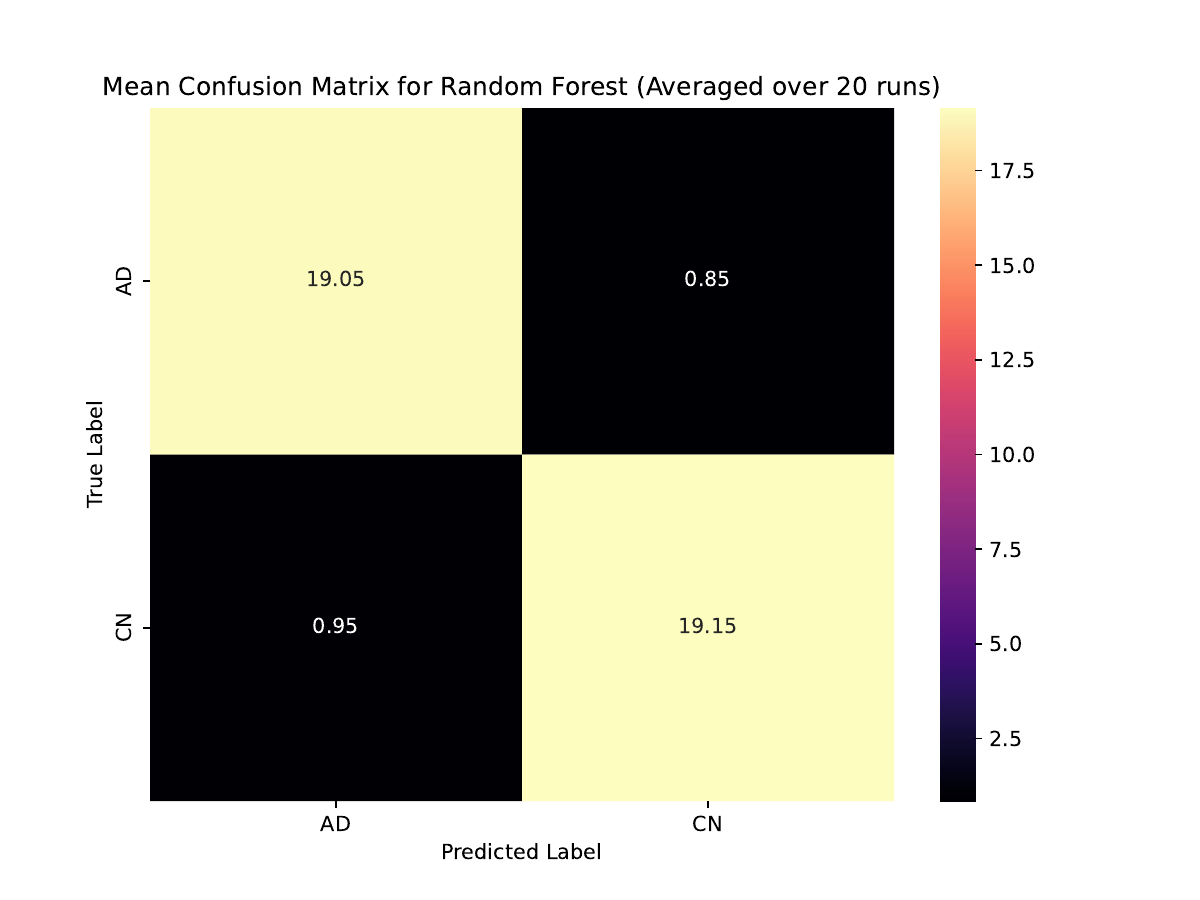}
    \caption{Random Forest}    
    \end{subfigure}
     \hspace{1.3em}
    \begin{subfigure}{0.3\textwidth}
    \includegraphics[width=\textwidth]{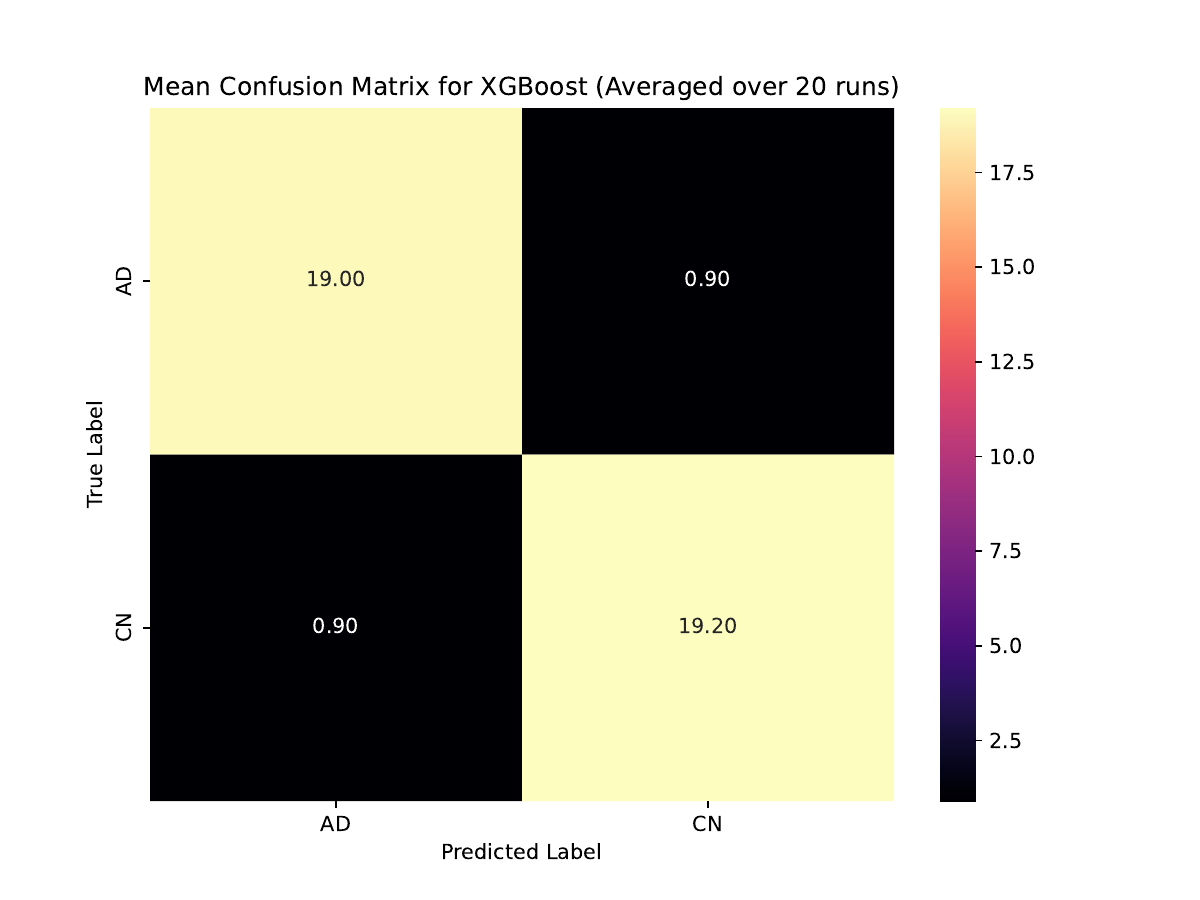}
    \caption{XGBoost}    
    \end{subfigure}
     \hspace{1.3em}
    \begin{subfigure}{0.3\textwidth}
    \includegraphics[width=\textwidth]{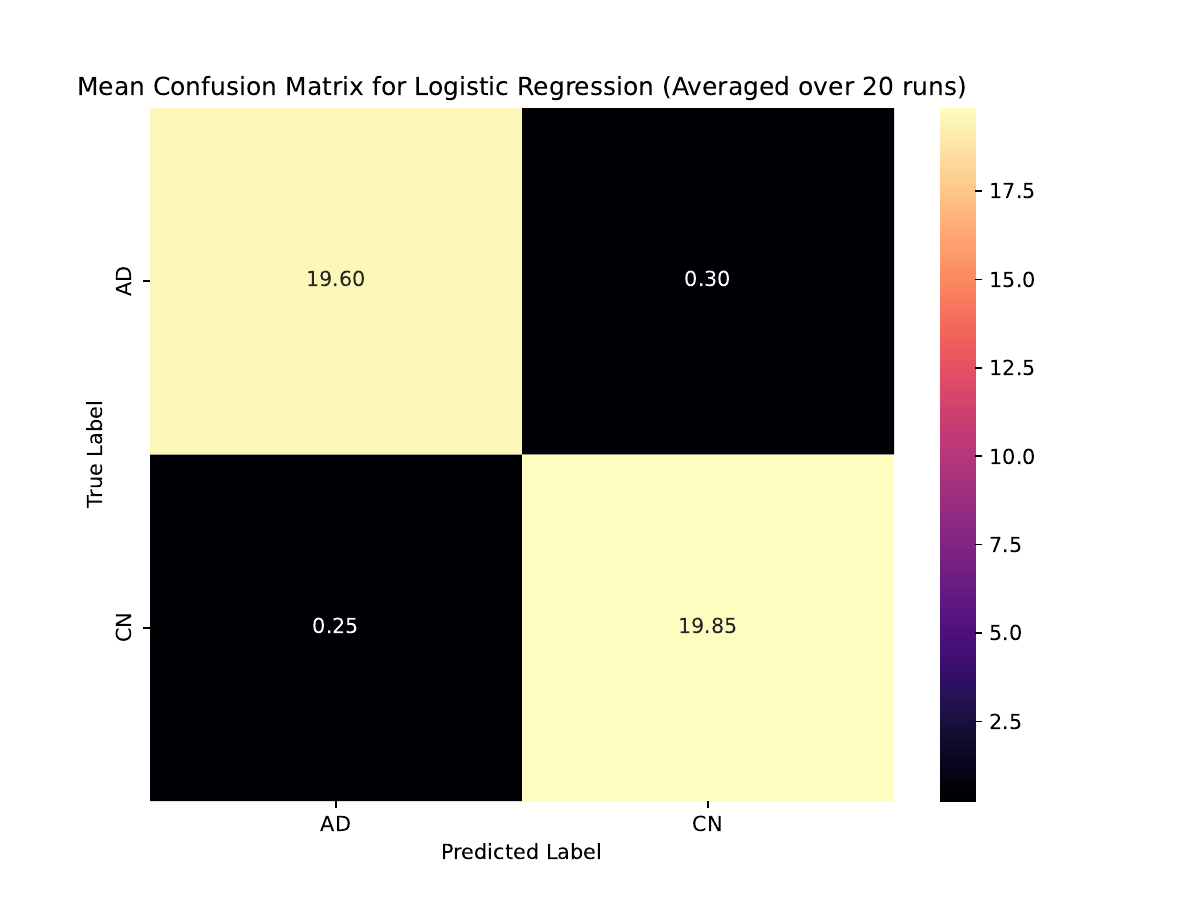}
    \caption{Logistic regression}    
    \end{subfigure}
    \begin{subfigure}{0.3\textwidth}
    \includegraphics[width=\textwidth]{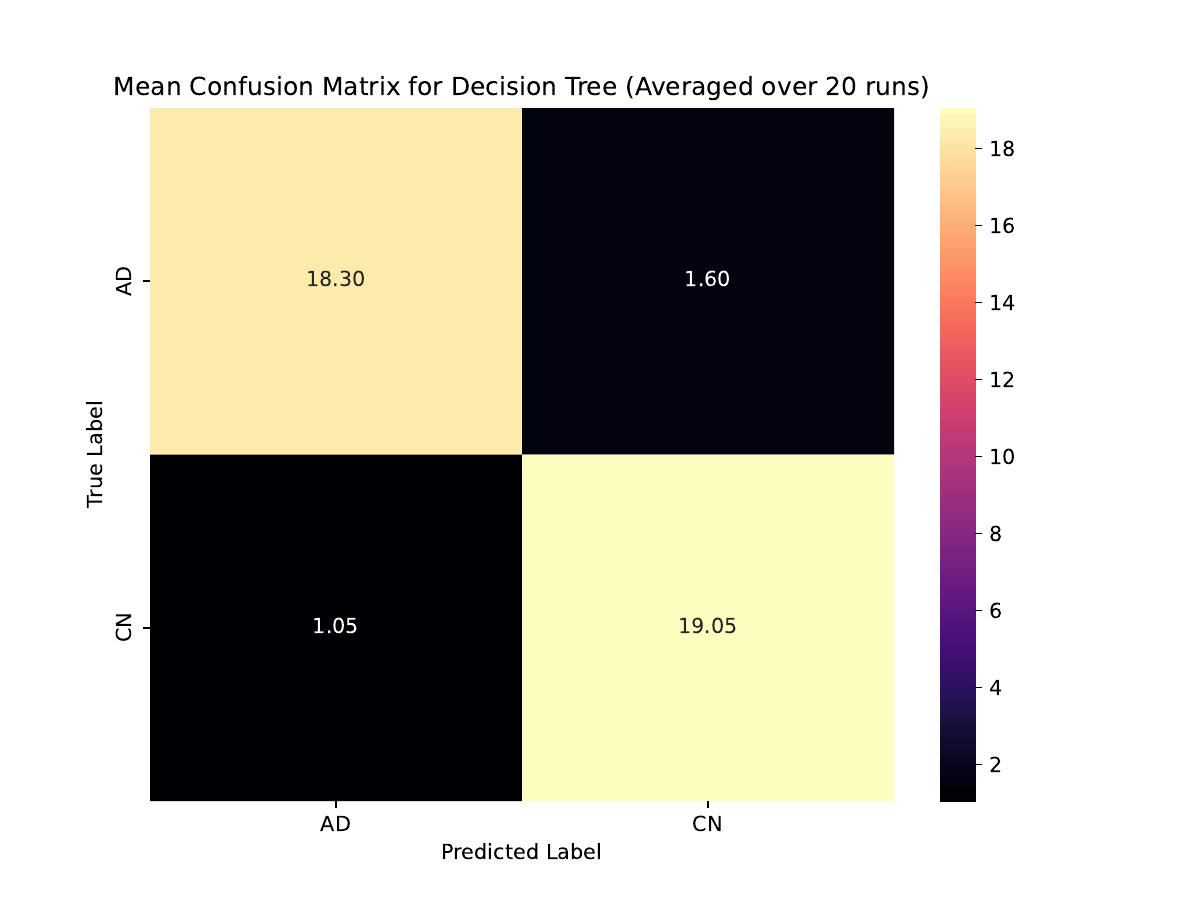}
    \caption{Decision tree}    
    \end{subfigure}
    \caption{The comparison of the seven classifiers in the confusion matrix.}
    \label{fig:confusionMats}
\end{figure}

\begin{figure}[!h]
    \centering
    \begin{subfigure}{0.48\textwidth}
    \includegraphics[width=\textwidth]{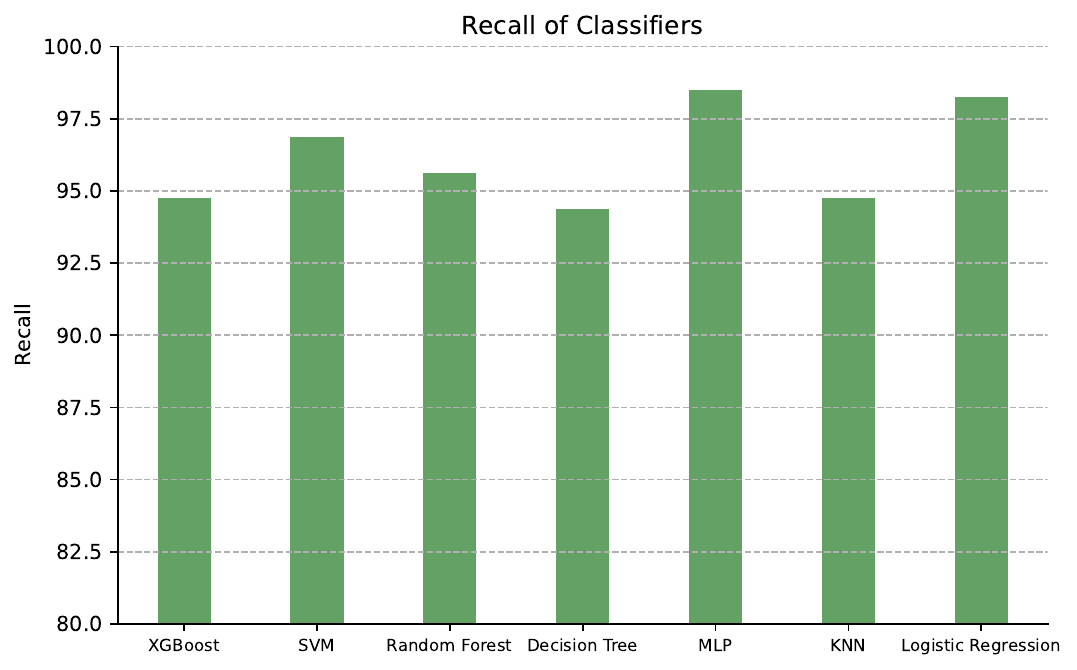}
    \caption{Recall}    
    \end{subfigure}
    \begin{subfigure}{0.48\textwidth}    
    \includegraphics[width=\textwidth]{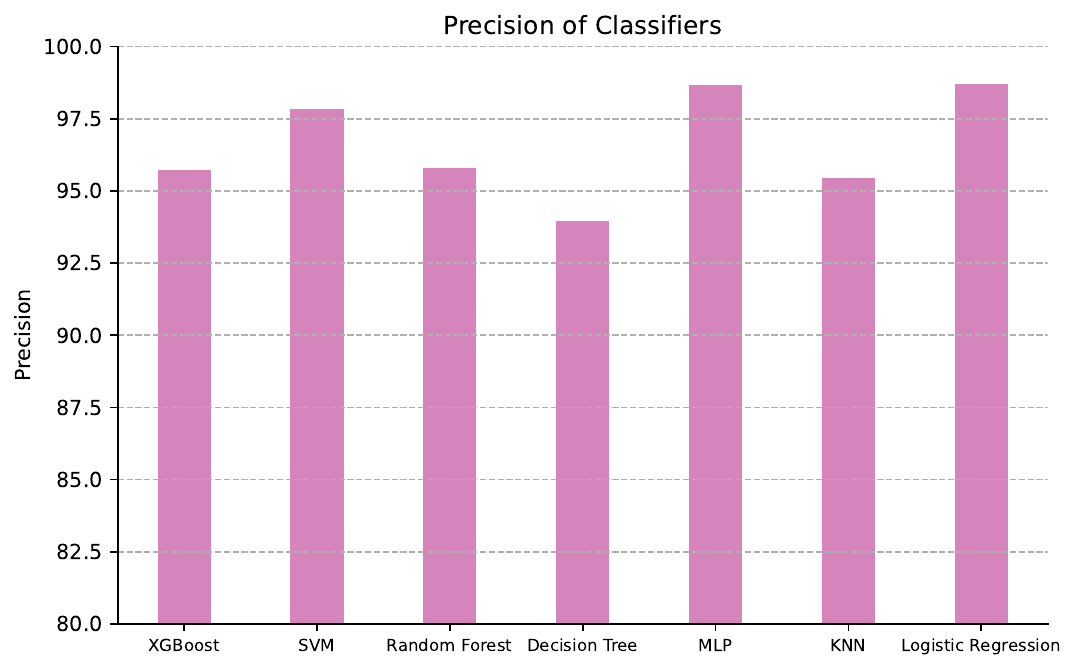}
    \caption{Precision}    
    \end{subfigure}
    \begin{subfigure}{0.48\textwidth}    
    \includegraphics[width=\textwidth]{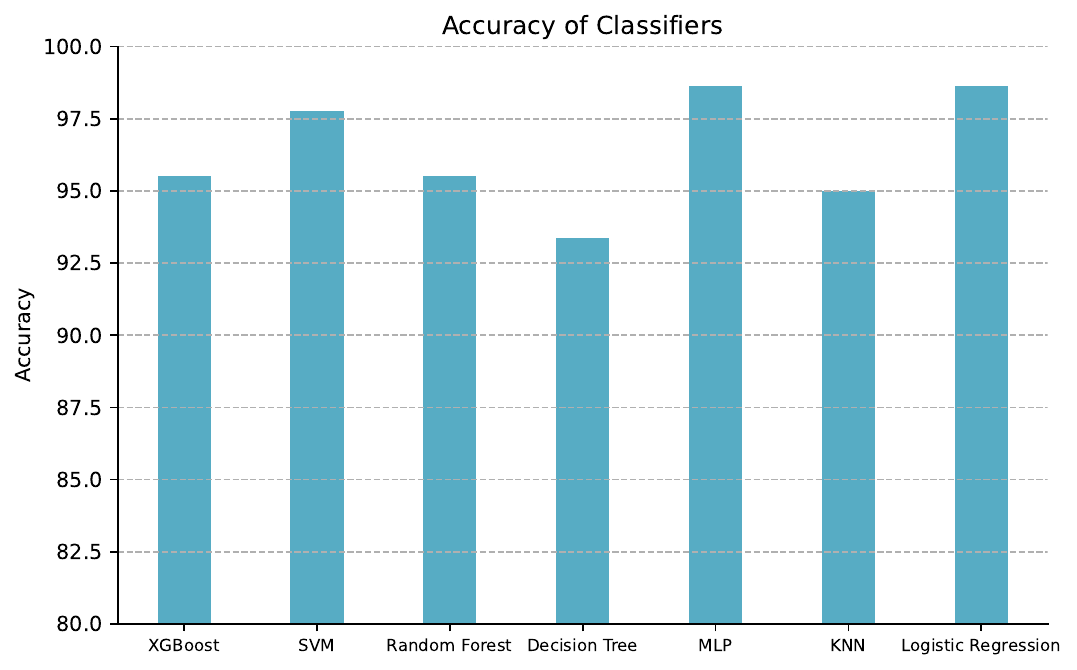}
    \caption{Accuracy}    
    \end{subfigure} 
    \begin{subfigure}{0.48\textwidth}
    \includegraphics[width=\textwidth]{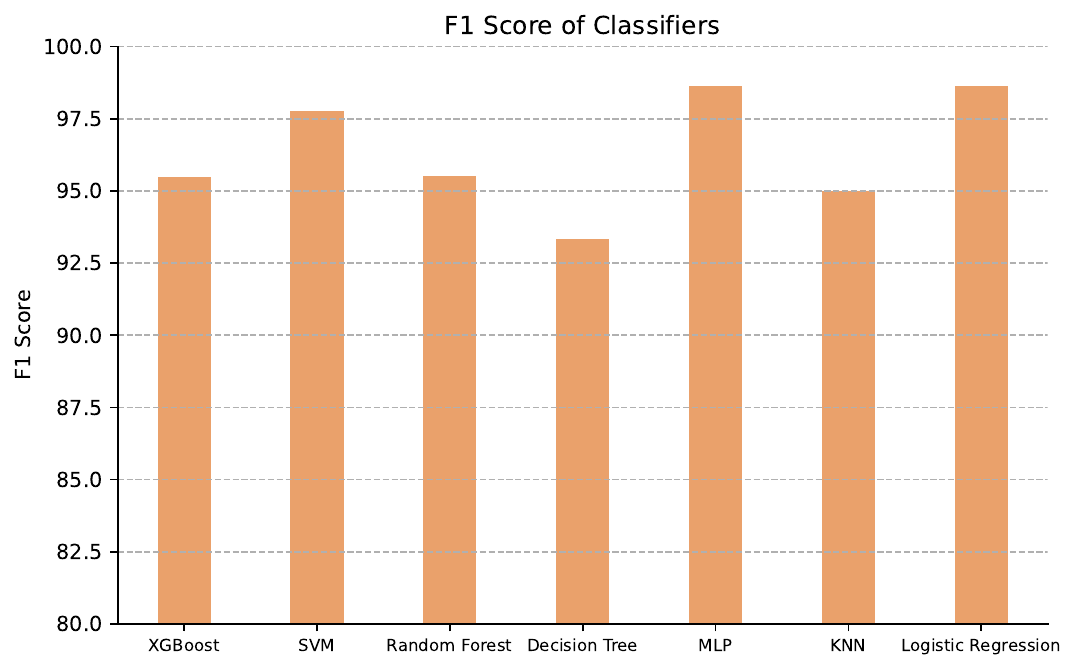}
    \caption{F$_1$ Score}    
    \end{subfigure}  
    \caption{The comparison of the seven classifiers using the classification measures.}
    \label{fig:MeasuresBinPlot}    
\end{figure}

In the classification phase, we utilized various classifiers, including XGBoost, SVM, Random Forest, MLP, Decision Tree, KNN, and logistic regression. We assessed the proposed signatures using an ADNI dataset that included 160 participants, consisting of 80 with Alzheimer's disease (AD) and 80 cognitively normal (CN). For the purposes of classification, we designate 80\% of the dataset as training samples and 20\% as test samples. By randomly selecting the training set and computing the average accuracy over ten iterations, we were able to obtain reliable results. Table \ref{tbl:classify}, Figures \ref{fig:confusionMats}, and \ref{fig:MeasuresBinPlot} show the result of the experiments. 

The confusion matrix is a performance evaluation tool commonly used in image classification. It is a square matrix that compares the actual class labels (ground truth) with the predicted class labels. Each row represents the true label or actual class, while each column represents the predicted label (class).

From the confusion matrix, important metrics such as sensitivity (recall), specificity, precision, accuracy, and F$_1$ Score can be derived. These measures presented in equations \eqref{Acc} to \eqref{F1}, offering insights into the model's performance for individual classes. This makes the confusion matrix a valuable tool for error analysis and identifying areas for improvement in classification models.
\begin{itemize}
\item \textbf{Accuracy:} Accuracy is the ratio of correctly predicted instances to the total instances in the dataset.  It indicates how often the classifier is correct overall.
\begin{equation}\label{Acc}
Accuracy = \dfrac{TP+TN}{TP+TN+FP+FN}
\end{equation}

\item \textbf{Sensitivity (Recall or True Positive Rate):} 
Sensitivity measures the proportion of actual positives that are correctly identified. It shows how well the model can identify positive cases. A high sensitivity indicates fewer false negatives.
\begin{equation}\label{Sen}
Sensitivity = \dfrac{TP}{TP+FN}
\end{equation}

\item \textbf{Specificity (True Negative Rate):} Specificity measures the proportion of actual negatives that are correctly identified. This metric indicates how well your model can identify negative cases, showing fewer false positives when it’s high. 
\begin{equation}\label{Spe}
Specificity = \dfrac{TN}{TN+FP}
\end{equation}

\item \textbf{Precision (Positive Predictive Value):} Precision quantifies how many selected items are relevant or true positives among all predicted positives.  High precision means that more positively classified images are indeed positive; it emphasizes reducing false positives in classifications. 
\begin{equation}\label{Pre}
Precision = \dfrac{TP}{TP+FP}
\end{equation}

\item \textbf{F$_1$ Score:} The F$_1$ Score is a harmonic mean between precision and recall, offering a balance between them, especially when there exists an uneven class distribution. The F$_1$ Score is particularly useful when you need to find an optimal balance between precision and recall. It is a better measure than accuracy in cases of imbalanced datasets, because it takes both false positives and false negatives into account.
\begin{equation}\label{F1}
F_1 Score = \dfrac{2*TP}{2*TP+FP+FN}
\end{equation}
\end{itemize}

The results indicate that the highest performance was attained by the MLP and Logistic Regression classifiers, with the experiment yielding a mean accuracy of 98.62\%. In the second position, SVM classifier attained an average accuracy of 96.88\% in the experiment. As already mentioned, the number of bins has an important impact on the result. By changing bin numbers, we chose four scales named Scale 1 to Scale 3 (see Table \ref{tbl:classify}).   
\begin{table}[!h]\tiny
	\caption{Performance comparison of machine learning classifiers across evaluation metrics. Values represent mean $\pm$ standard deviation computed over 20 independent train-test splits. Best performing values for each metric are highlighted in bold.}
	\begin{center}
		\begin{tabular}{lllllll}		
			\hline			
			\multicolumn{1}{l}{{\textbf{Classifier}}} &
			\multicolumn{1}{l}{{\textbf{Scale}}} &
			\multicolumn{1}{l}{{\textbf{Accuracy}}} &
			\multicolumn{1}{l}{{\textbf{Precision }}} &
			\multicolumn{1}{l}{{\textbf{Recall}}} &
			\multicolumn{1}{l}{{\textbf{$F_1 Score $ }}} \\				
			& & (Mean ± Std) &  (Mean ± Std) &  (Mean ± Std) &  (Mean ± Std)\\ 	
			\hline
			\textbf{XGBoost} & Scale 1 & 79.88 $\pm$  4.64 & 	81.1 $\pm$ 	4.88  &	79.88 $\pm$	4.64 &	79.74 $\pm$	4.71 \\
			& Scale 2 & 91.88 $\pm$	3.86 &	92.44 $\pm$	3.64 &	91.88 $\pm$	3.86 &	91.87	$\pm$ 3.86\\
			& Scale 3 & 95.5	 $\pm$3.84	& 95.71	 $\pm$ 3.74 &	95.5  $\pm$	3.84 &	95.5  $\pm$	3.85  \\		
			\hline			
			\textbf{Random Forest} & Scale 1 &	80.12 $\pm$ 5.67	& 81.68  $\pm$	5.33	  & 80.12 $\pm$ 5.67	& 79.94  $\pm$	5.83
 \\			
			& Scale 2  &	93.38 $\pm$	3.38	 & 93.71 $\pm$	3.3	& 93.38 $\pm$	3.38	 & 93.38 $\pm$	3.38 \\
			& Scale 3 & 95.5  $\pm$	4.3	& 95.79  $\pm$	4.08 &	95.5	  $\pm$ 4.3	& 95.51	 $\pm$ 4.29
 \\			
			\hline
			\textbf{SVM} & Scale 1  & 80.25 $\pm$ 5.86	 & 81.78  $\pm$ 5.3	 & 80.25  $\pm$ 5.86	& 80.06	$\pm$ 6.01 \\
			& Scale 2  &	93.25 $\pm$	3.8	& 93.64 $\pm$	3.65 &	93.25 $\pm$	3.8 & 	93.26 $\pm$	3.79 \\
			& Scale 3  &	97.75  $\pm$	2.49 &	97.85  $\pm$	2.4  & 97.75  $\pm$	 2.49	 & 97.75  $\pm$	2.49\\
			\hline				
			\textbf{Decision Tree} & Scale 1 & 75.62 $\pm$	6.56 & 	76.38 $\pm$	6.37	  &  75.62 $\pm$	6.56 &	75.57	$\pm$ 6.56\\
				& Scale 2  & 90.75 $\pm$	3.72	 & 91.15	 $\pm$ 3.59 &	90.75 $\pm$	3.72	 & 90.76	 $\pm$ 3.71\\
				& Scale 3  & 93.38  $\pm$	4.28 &	93.96  $\pm$	3.8 & 93.38  $\pm$ 	4.28 &	93.34  $\pm$ 	4.32 \\
				\hline									
			\textbf{MLP} & Scale 1 & 79.12 $\pm$	4.63 & 	80.24	$\pm$4.6 &	79.12 $\pm$	4.63	 & 78.99 $\pm$	4.66\\
				& Scale 2  &93.88	$\pm$2.68	 & 94.31 $\pm$	2.45 &	93.88 $\pm$	2.68	 & 93.87 $\pm$	2.68 \\				
				& Scale 3  & \textbf{98.62  $\pm$1.47} &	\textbf{98.68  $\pm$	1.43} & \textbf{98.62  $\pm$	1.47} & \textbf{98.62  $\pm$ 1.47}\\
				\hline										
			\textbf{KNN} & Scale 1  & 78 $\pm$	5.57 & 	79.39 $\pm$ 	5.01	 & 78 $\pm$	5.57 & 	77.77 $\pm$	5.77 \\
				& Scale 2 & 91.5 $\pm$	4.06	 &92.25	$\pm$ 3.43 &	91.5 $\pm$	4.06	 & 91.47 $\pm$	4.09 \\			
				& Scale 3 & 95	 $\pm$ 3.54 &	95.46  $\pm$	2.94 & 95  $\pm$	3.54 &	94.99	 $\pm$ 3.55 \\
				\hline				
			\textbf{Logistic Regression} & Scale 1  & 79 $\pm$ 5.78 & 	80.08	$\pm$ 5.96  &	79	$\pm$ 5.78 & 	78.93 $\pm$	5.81
\\
				& Scale 2  &94.38 $\pm$	2.48 &	94.85 $\pm$	2.26& 94.38 $\pm$	2.48	 & 94.39 $\pm$	2.48\\
				& Scale 3 & \textbf{98.62  $\pm$	2.68} & \textbf{	98.7   $\pm$ 2.54} & \textbf{98.62  $\pm$	2.68	}  & \textbf{98.62  $\pm$ 	2.68} \\
				\hline							
		\end{tabular}
	\end{center}
	\label{tbl:classify}
\end{table}

To assess the statistical significance of differences in classifier accuracy, we performed Welch’s t-test, which accounts for unequal variances between groups. The results (see Table \ref{ttestTable}) show that all classifiers except Logistic Regression exhibit significantly lower accuracy than the MLP baseline ($p < 0.05$). Notably, Logistic Regression achieved accuracy statistically indistinguishable from MLP, because it has the same accuracy as MLP. For all other methods, the significantly lower accuracy underscores the effectiveness of MLP as a strong baseline in this task.
\begin{table}[!h]\tiny
\caption{Statistical comparison of classifier accuracy against MLP baseline using Welch's t-test. All classifiers except Logistic Regression show significantly lower accuracy than MLP (p < 0.05).}
\begin{center}
\begin{tabular}{lcccc}
\hline
\textbf{Model} & \textbf{Accuracy (Mean ± Std)} & \textbf{t-statistic (vs. MLP)} & \textbf{p-value} & \textbf{Significance} \\
\hline
MLP (Baseline) & 98.62  $\pm$	1.47 & - & - & -\\
XGBoost &  95.5	 $\pm$ 3.84 & 2.91 & 0.0062 & sig. \\
SVM & 97.75  $\pm$	2.49 & 2.30  & 0.026 & sig. \\
Random Forest & 95.5  $\pm$	4.3 & 3.57 & 0.001 & sig. \\
Decision Tree & 93.38  $\pm$	4.28 & 4.89 & 1.996e-05  & sig. \\
KNN & 95	 $\pm$ 3.54 & 5.53 & 4.26e-06 & sig. \\
Logistic Regression & 98.62  $\pm$	2.68 & - & - & -\\
\hline
\end{tabular}
\end{center}
\label{ttestTable}
\end{table}


\begin{figure}
    \begin{subfigure}{0.45\textwidth}
    \includegraphics[width=\textwidth]{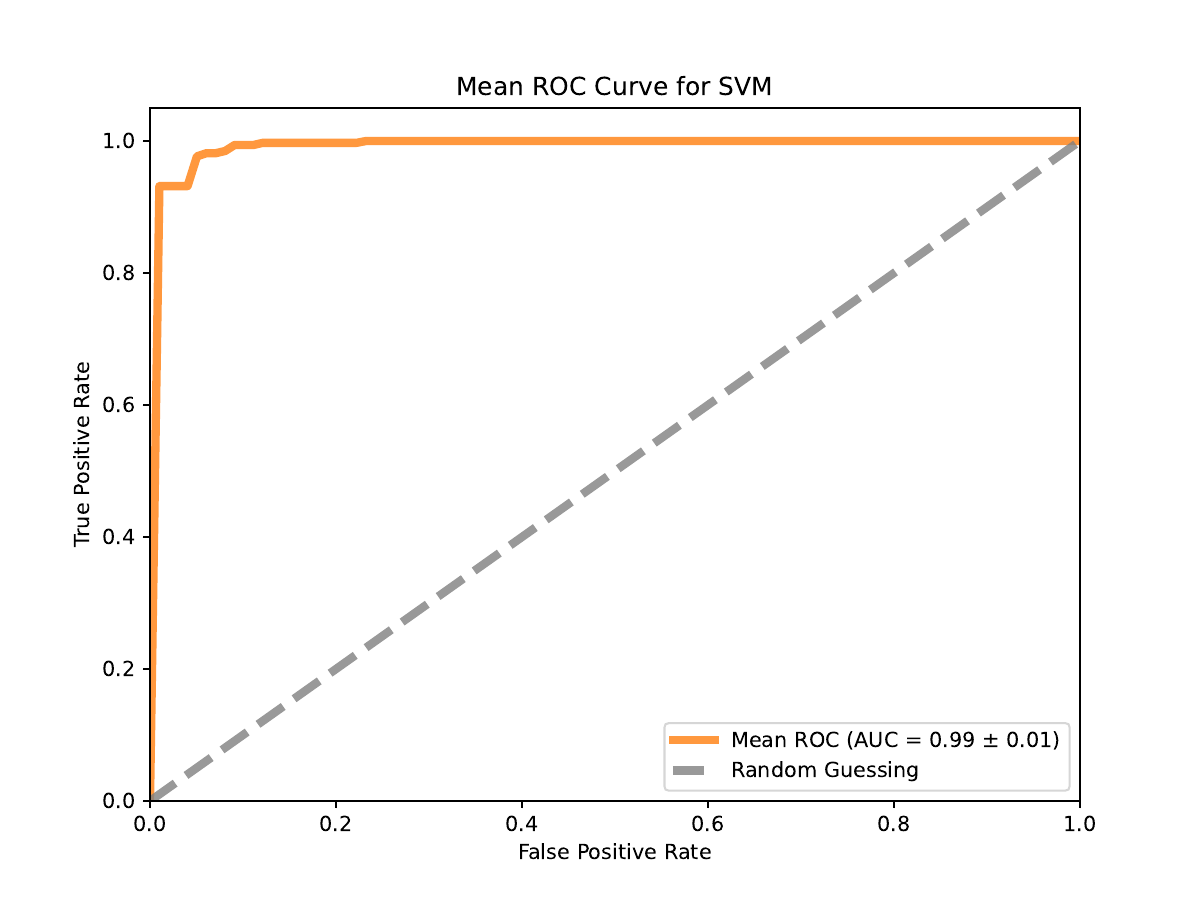}
    \caption{}    
    \end{subfigure}
    \hspace{0.5em}
    \begin{subfigure}{0.45\textwidth}
    \includegraphics[width=\textwidth]{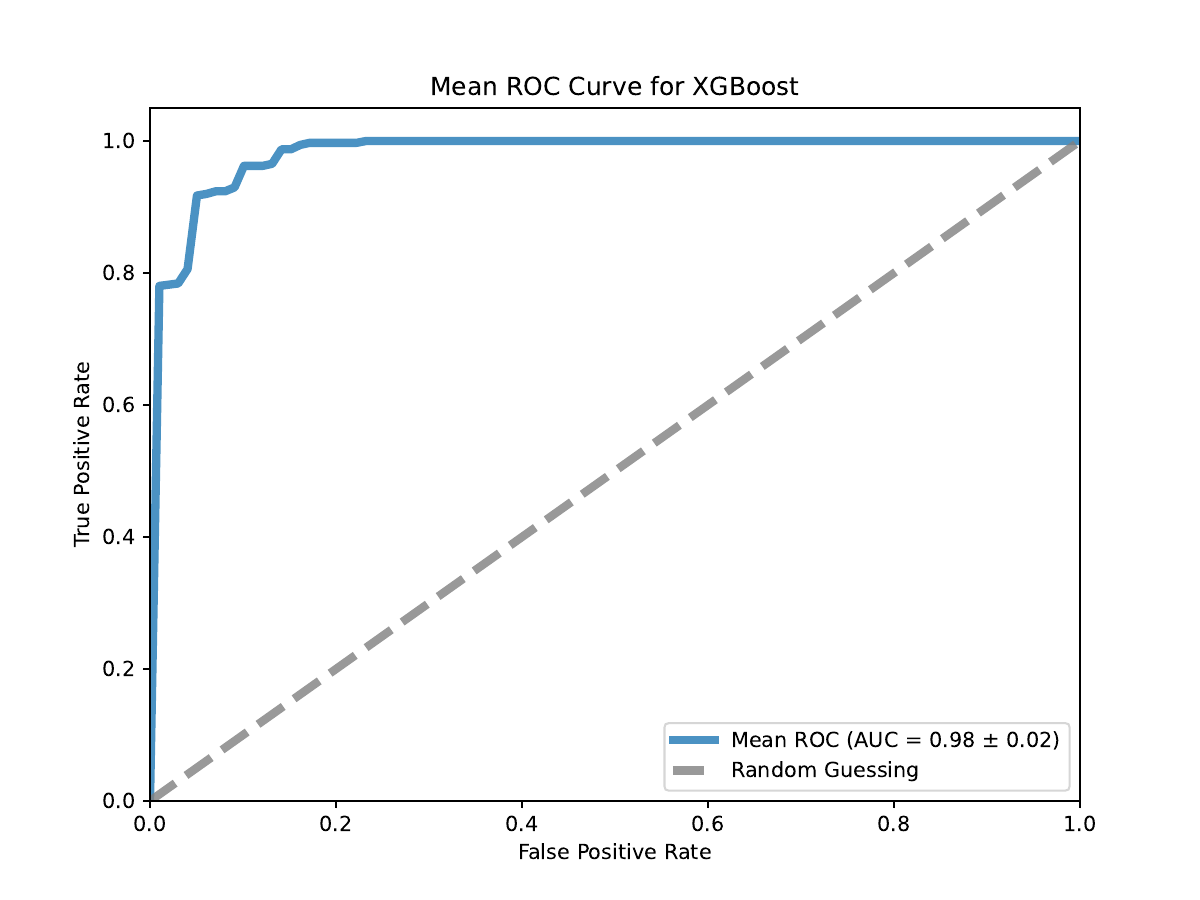}
    \caption{}    
    \end{subfigure}
     \hspace{0.5em}
    \begin{subfigure}{0.45\textwidth}
    \includegraphics[width=\textwidth]{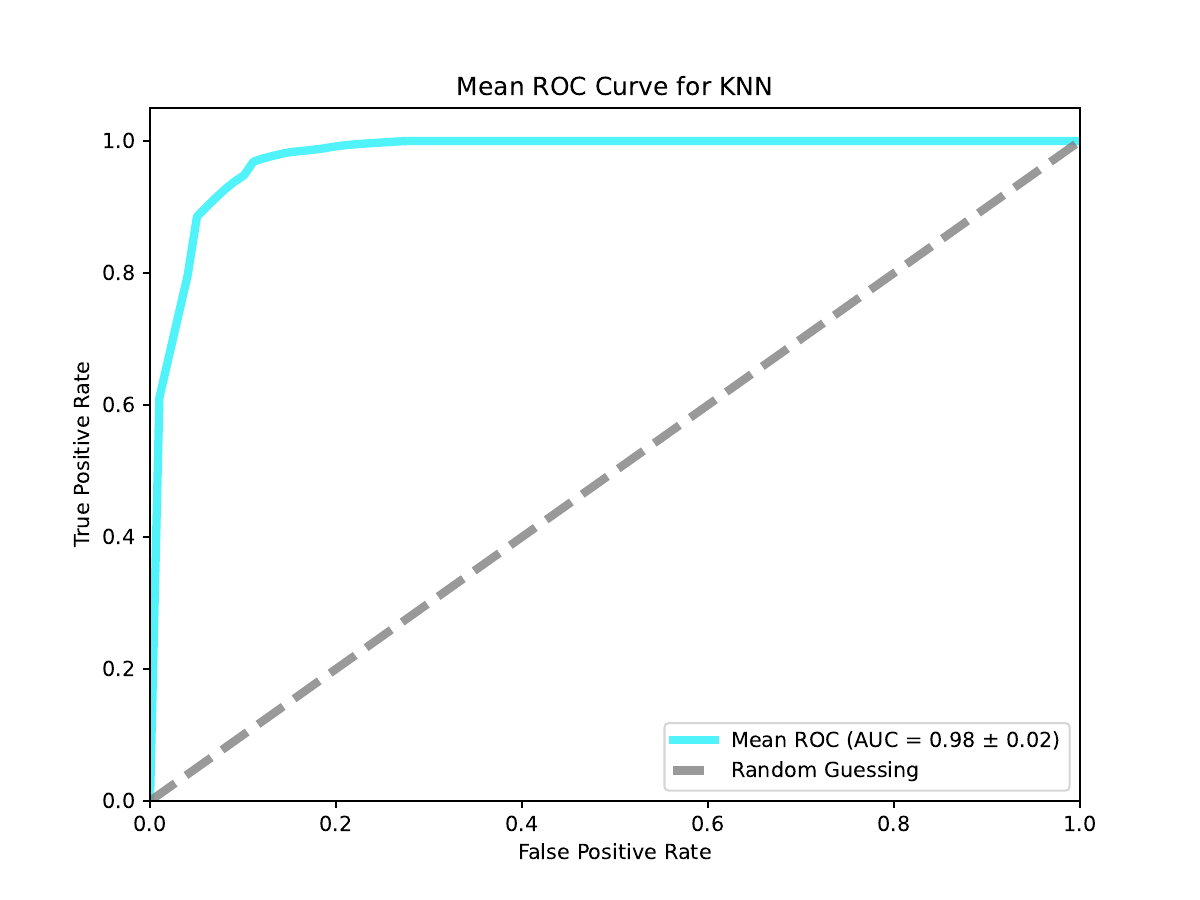}
    \caption{}    
    \end{subfigure}
     \hspace{1.6em}
    \begin{subfigure}{0.45\textwidth}
    \includegraphics[width=\textwidth]{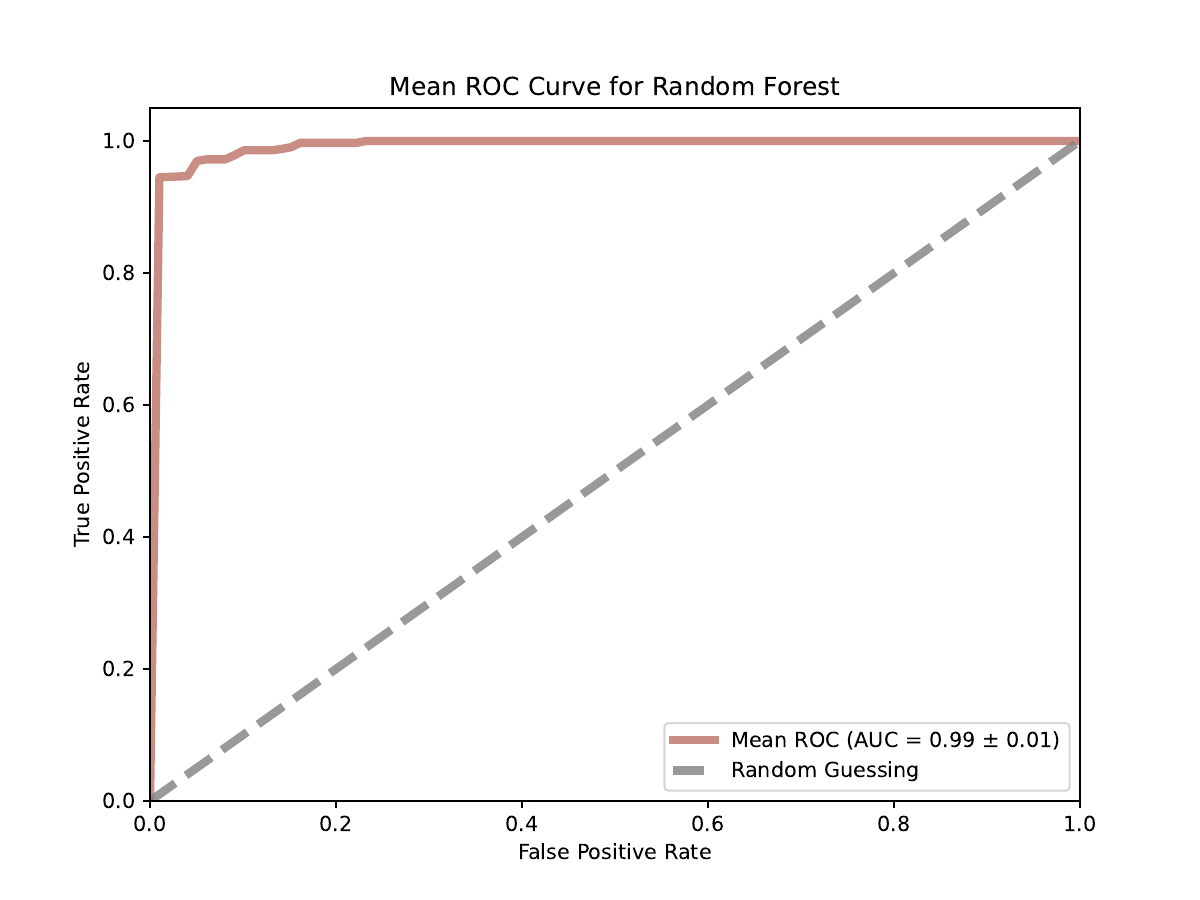}
    \caption{}    
    \end{subfigure}
     \hspace{1.9em}
     \begin{subfigure}{0.45\textwidth}
    \includegraphics[width=\textwidth]{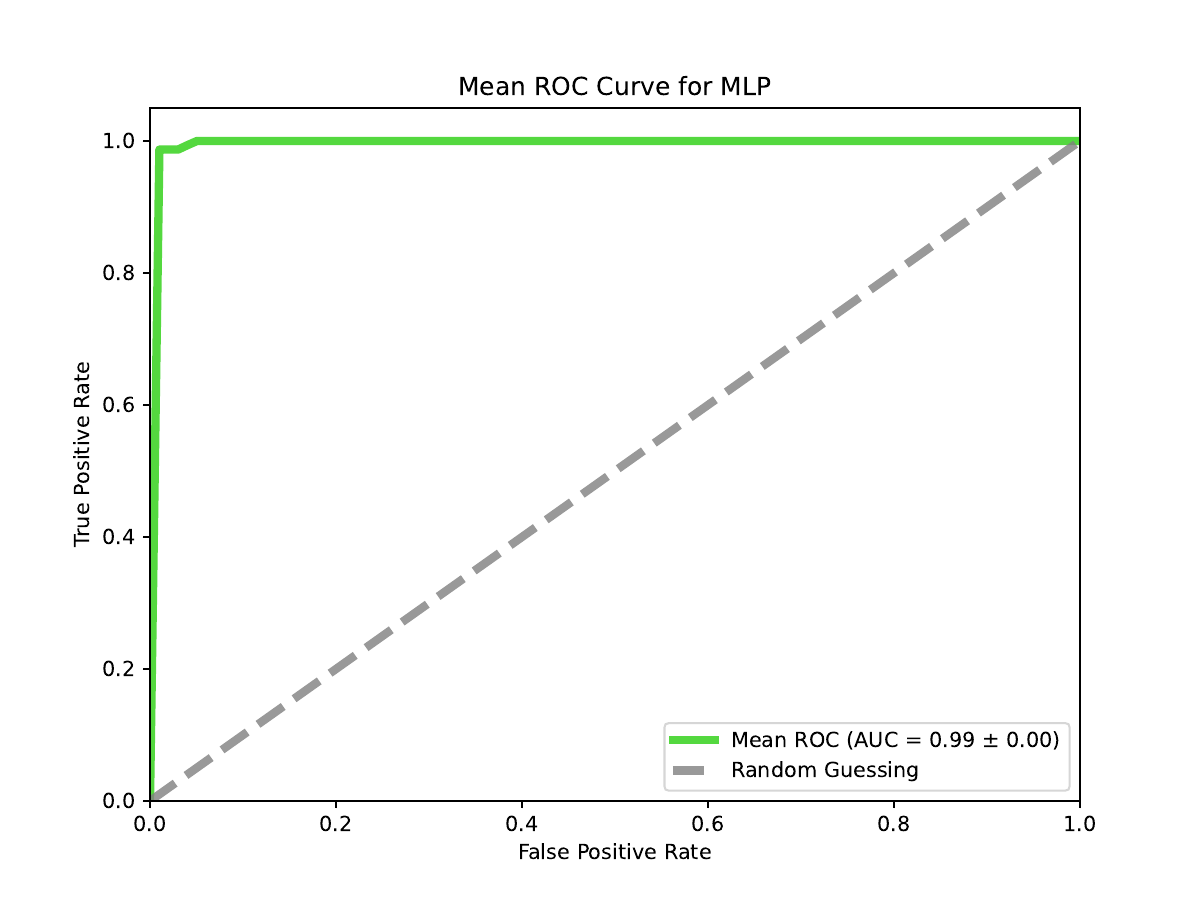}
    \caption{}    
    \end{subfigure}
     \hspace{2.5em}
    \begin{subfigure}{0.45\textwidth}
    \includegraphics[width=\textwidth]{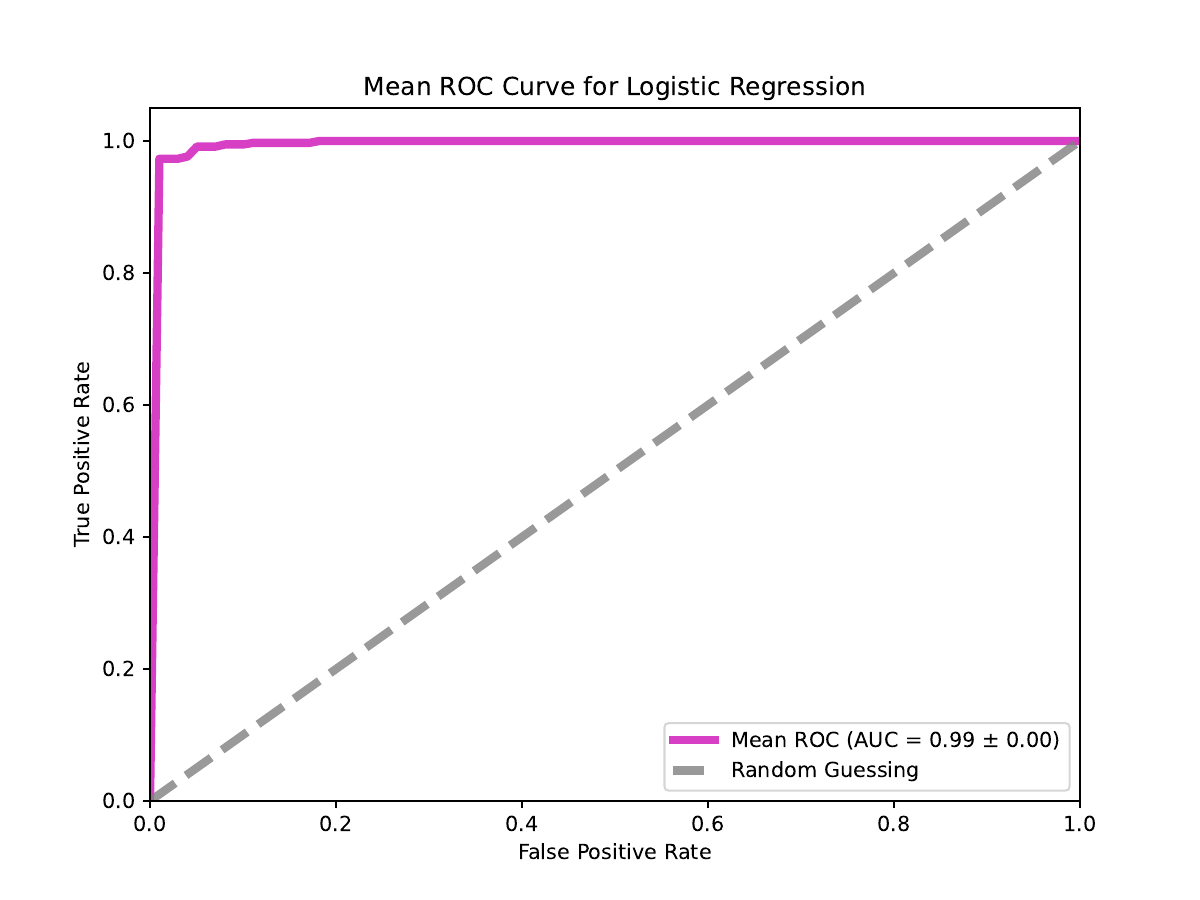}
    \caption{}    
    \end{subfigure}
    \begin{subfigure}{0.45\textwidth}
    \includegraphics[width=\textwidth]{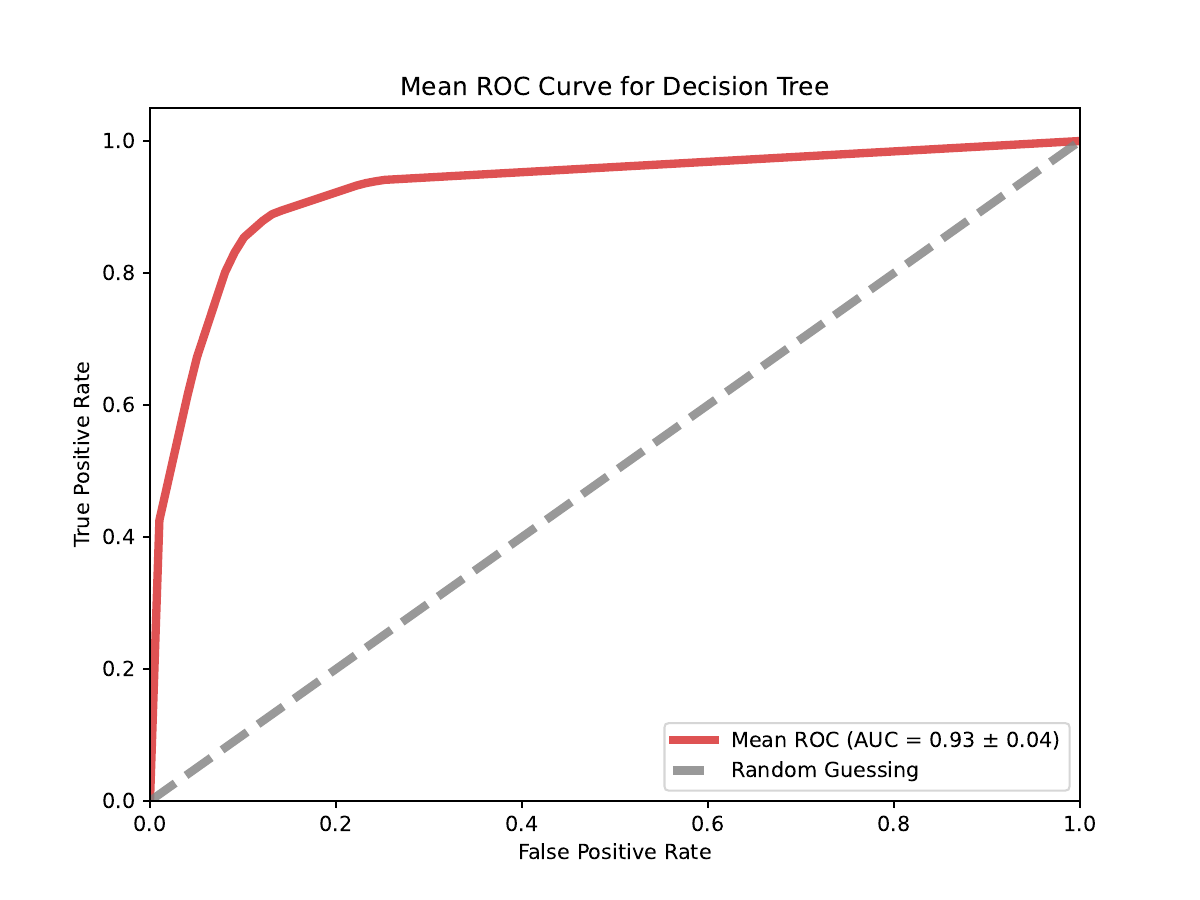}
    \caption{}    
    \end{subfigure}
    \caption{The ROC curve of classifiers in the best performance.}
    \label{fig:confusionMat}
\end{figure}
Advanced techniques were compared with ours in a comparison that can be observed in Table \ref{ComparisionTbl}.  
\begin{flushleft}
\begin{table}[!h]\tiny
	\caption{Comparison of the proposed methods with the state-of-the-art methods on the ADNI dataset.}	
	\begin{center}			
		\begin{tabular}{lllll}			
		\hline			
		\multicolumn{1}{l}{\textbf{Author}} &
		\multicolumn{1}{l}{\textbf{Year}} &
		\multicolumn{1}{l}{\textbf{Method}} &
		\multicolumn{1}{l}{\textbf{Accuracy}} \\
		\hline
		Zeng et al. \citep{zeng2013teichmuller} & 2013 & Teichmüller shape descriptor + SVM & 91.38 \\	\\			
		Shi et al. \citep{shi2019hyperbolic} & 2019 & hyperbolic Wasserstein distance & 76.7 \\\\
		Razib \citep{razib2017structural} & 2017 & Tutte's embedding and Harmonic mapping + KNN & 88 \\\\
		Acharya et al. \citep{acharya2019automated} & 2019 &	Shearlet transform + KNN & 94.54 \\\\
		Qin et al. \citep{qin20223d} & 2022 & 3D Residual U-Net model with hybrid attention technique & 92.68\\\\
		Zhang et al. \citep{zhang2022diagnosis} & 2022 & sMRI gray matter segments + deep learning & 90 \\\\
		Kushol et al. \citep{kushol2022addformer} &	2022 & vision Transformer &	88.20 \\\\
		Kong et al. \citep{kong2022multi}	& 2022 &  fuse MRI with PET images + 3D CNN  & 93.21 \\\\
            Zhang et al. \citep{zhang2023multi} & 2023 & multi-modal (sMRI+FDG-PET+CSF) cross-attention & 91.07 \\\\
            Abbas et al. \citep{abbas2023transformed} & 2023 & Jacobian map + CNN & 96.61 \\ \\
            Ahmadi et al. \citep{AHMADI2024106212} & 2024 & covariance-based descriptors in Ricci energy optimization +   &  \\
            & & manifold-based classification using KNN & 96 \\\\
            Ahmadi et al. \citep{ahmadi2024alzheimer} & 2024 & conformal descriptors + XGBoost & 96.88 \\ \\
		Proposed Method & - &area distortion, conformal factor, and Gaussian curvature descriptors +  &  \\
		& & MLP (XGBoost, RF, SVM, KNN, DT, LR) & $\textbf{98.62}$\\
		\hline				
		\end{tabular}	
		\end{center}
	\label{ComparisionTbl}
\end{table}
\end{flushleft}
\section{Discussion}
This study introduced a novel, landmark-free computational pipeline for distinguishing between Alzheimer's disease (AD) and cognitively normal (CN) individuals based on the conformal geometry of the hippocampal surface. Our method leveraged Euclidean Ricci flow for planar parameterization to extract robust geometric features—conformal factor, area distortion, and intrinsic Gaussian curvature—which were then encoded into powerful statistical descriptors using Shannon entropy. The central finding of this work is that this approach achieved exceptional classification performance, with Multi-Layer Perceptron (MLP) and Logistic Regression models reaching a mean accuracy of 98.62\% on a balanced dataset from the ADNI cohort. This performance not only validates the efficacy of our proposed features but also positions our method as a state-of-the-art technique for computational neuroanatomy in AD diagnosis.

The superior performance of our method can be attributed to the synergistic combination of theoretically grounded geometric features and effective feature encoding. The conformal factor directly measures the local scaling required to achieve a conformal map, sensitively capturing subtle expansions or contractions of the cortical surface caused by AD-related atrophy. Area distortion, a natural byproduct of this process, quantifies the deviation from isometry, providing a complementary measure of morphological change that is more pronounced in diseased tissue. Crucially, by calculating Gaussian curvature on the original mesh prior to parameterization, we captured intrinsic geometric properties that are preserved and informative of the underlying neuroanatomical integrity, which may be normalized during the conformal mapping process. The application of Shannon entropy to these feature maps was a critical step, successfully transforming complex, high-dimensional spatial distributions into single, highly discriminative scalars for each subject. This encoding effectively summarizes the overall "geometric disorder" or loss of structural organization characteristic of AD pathology.

Our results place this work firmly within the context of existing literature while highlighting its unique contributions. The achieved accuracy of 98.62\% surpasses that of many other advanced methods, including deep learning approaches on neuroimaging data (e.g., Qin et al., 92.68\% \cite{qin20223d}; Kong et al., 93.21\% \cite{kong2022multi}), other surface-based conformal methods (e.g., Zeng et al., 91.38\% \cite{zeng2013teichmuller}; Shi et al., 76.7\% \cite{shi2019hyperbolic}), and our own previous work (96.88\% in \cite{ahmadi2024alzheimer} and 96\% in \cite{AHMADI2024106212}). This improvement underscores a key advantage of our pipeline: its landmark-free nature. By eliminating the need for manual landmarking—a process that is time-consuming, expert-dependent, and difficult to scale—our method offers superior scalability and accessibility for large-scale clinical studies and potential automated diagnostic applications. Furthermore, the strong performance of simpler models like Logistic Regression, matching that of the more complex MLP, suggests that our entropy-encoded features are inherently highly separable, providing a clean and robust signal for classification.

The choice of the hippocampal region was well-justified, as it is one of the earliest and most severely affected structures in AD. Focusing our analysis here likely increased the signal-to-noise ratio compared to whole-brain analyses, allowing our geometric descriptors to pinpoint disease-specific changes with high sensitivity. The finding that performance improved across "Scale 1" to "Scale 3" (from ~80\% to >98\% accuracy) indicates that the binning strategy for entropy calculation is crucial. This likely reflects an optimization in capturing the most discriminative statistical distribution of the geometric features, with finer scales (Scale 3) providing a more detailed and informative representation of the underlying pathology.

Despite these compelling results, several limitations should be acknowledged. First, the study was conducted on a dataset of 160 subjects from the ADNI database. While this provides a solid proof-of-concept, validation on larger, more diverse, and multi-site datasets is essential to confirm the generalizability of our findings and mitigate potential biases. Second, our study focused on a binary classification task (AD vs. CN). The model's performance in distinguishing AD from other dementias or in predicting progression from Mild Cognitive Impairment (MCI) to AD remains to be investigated and represents a critical future direction. Third, while planar parameterization was effective, its comparative performance against spherical Ricci flow parameterization in this specific context could be explored. Finally, the clinical translation of such a tool would require integration into radiologists' workflows and validation in real-time clinical settings.

Future work will focus on several avenues. Firstly, we plan to apply this pipeline to larger, publicly available datasets and to multi-class problems, including MCI converters and non-converters. Secondly, exploring different feature encoding strategies, such as deep learning autoencoders or other information-theoretic measures, could potentially yield further improvements. Thirdly, extending the analysis to a longitudinal framework to track geometric changes over time could provide valuable biomarkers for monitoring disease progression and treatment efficacy. Finally, investigating the applicability of this method to other neurological and psychiatric disorders characterized by cortical atrophy, such as schizophrenia or epilepsy, would test the generality of our approach.

In conclusion, we have presented a highly accurate, automated, and landmark-free framework for the classification of Alzheimer's disease based on hippocampal surface geometry. By leveraging the theoretical power of discrete Ricci flow and the practical utility of Shannon entropy, we have derived a set of robust features that capture the essence of AD-related neuroanatomical change. The exceptional performance of our method, which outperforms existing state-of-the-art techniques, highlights its significant potential as a powerful tool for computer-aided diagnosis and a valuable asset in the ongoing fight against Alzheimer's disease.
\section{Conclusion}\label{sec13}
This study investigates Alzheimer's disease by analyzing brain surface data through localized processing, integrating geometric and conformal features using Ricci flow parameterization and Shannon entropy for surface indexing. Multiple classifiers—including XGBoost, SVM, Decision Tree, Random Forest, MLP, KNN, and Logistic Regression—are utilized for robust classification.

During feature extraction, we introduce novel signatures area distortion entropy, conformal factor entropy, and Gaussian curvature entropy, computed from surface triangulation meshes. While area distortion and conformal factor are obtained via Ricci flow parameterization, Gaussian curvature is directly calculated from the primary surface. Evaluated on the ADNI dataset, our approach achieves a high accuracy of 98.62\% in differentiating healthy individuals from Alzheimer's patients. Beyond hippocampal shape analysis, this method can be extended to other surfaces, such as facial expression recognition.

Additionally, the framework shows promise for studying other neurological and psychiatric disorders, such as schizophrenia. Future work could further assess its precision and adaptability across broader applications. We anticipate that this methodology will enhance disease detection and diagnosis, ultimately advancing clinical care and patient outcomes.

\section*{Acknowledgments}
We would like to thank the Alzheimer’s Disease Neuroimaging Initiative (ADNI) for providing access to the data required for this study.

\section*{Declaration of competing interest}
The authors declare that they have no known competing financial interests or personal relationships that could have appeared to influence the work reported in this paper.

\section*{Appendix}
\begin{figure}[!h]
    \centering
    \begin{subfigure}{0.49\textwidth}
\includegraphics[width=\textwidth]{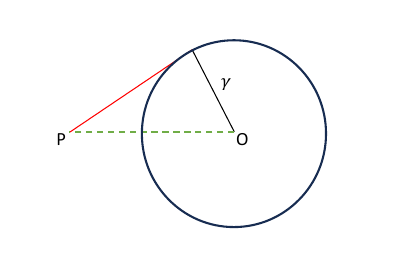}
\caption{}
\label{fig:Power}
\end{subfigure}
\begin{subfigure}{0.49\textwidth}
\includegraphics[width=\textwidth]{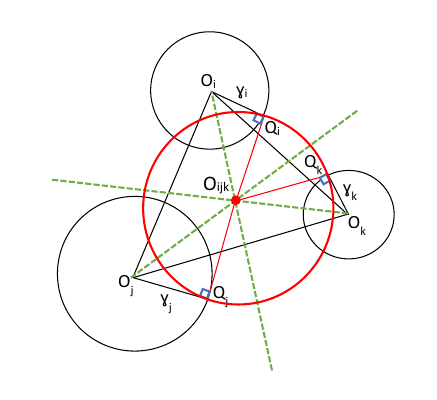}
\caption{}
\label{fig:PowerCenterofTriangle}
\end{subfigure}
    \caption{Power Circle.}
    \label{fig:PowerCenter}
\end{figure}

\textbf{Computing power center of a triangle:}\\
Given a triangle $f_{ijk}$ with vertices $ v_i $, $ v_j $, and $ v_k $. Construct circles centered at points $ v_i $, $ v_j $, and $ v_k $ with radii equal to distances from these points to another point (which could be any arbitrary distance). Lets denote these circles as: \\
Circle $C_i$ at vertex $ v_i $ centered at point $ O_i $ with radius $ \gamma_i $. Circle $C_j$ at vertex $ v_j $ centered at point $ O_j $ with radius $ \gamma_j $. Circle $C_k$ at vertex $ v_k $ centered at point $ O_k $ with radius $ \gamma_k $.\\
The power relative to circle $C$ centered at point $O$ with radius $\gamma$ can be calculated using (See Figure \ref{fig:Power}): 
\begin{equation}
P = {PO}^2 - \gamma^2,
\end{equation}
where $ PO $ is the Euclidean distance from $P$ to $O$ can be replaced with $\Vert P-O \Vert ^2$.\\
The concept relates closely to "the power of a point," which states that for any external point relative to these circles, there exists a relationship between distances from that external point to different parts of those circles. For computing power center of a triangle, we should find radical axis between each two circles located in the triangle (See Figure \ref{fig:PowerCenterofTriangle}, the radical axis is represented by a green dotted line). The radical axis of two circles is the locus of points that have equal power with respect to both circles, in other words the radical axis of two circles $C_i$ and $C_j$ is computed by $Power(P,Ci) = Power(P,Cj)$ that represent $\Vert P-O_i \Vert ^2 - \gamma_i^2 = \Vert P-O_j \Vert ^2 - \gamma_j^2$. Similarly, for the three circles $C_i$, $C_j$, and $C_k$, the following equation is formed: 
\begin{equation} \label{Eq:powerCenterofTrianle}
	\begin{cases} 
	\Vert P-O_i \Vert ^2 - \gamma_i^2 = \Vert P-O_j \Vert ^2 - \gamma_j^2 \\
	\Vert P-O_j \Vert ^2 - \gamma_j^2 = \Vert P-O_k \Vert ^2 - \gamma_k^2 \\
	\Vert P-O_i \Vert ^2 - \gamma_i^2 = \Vert P-O_k \Vert ^2 - \gamma_k^2
	\end{cases}
\end{equation}
By solving this system of equations, the power center $O_{ijk}$ of the triangle $f_{ijk}$ is calculated (See Figure \ref{fig:PowerCenterofTriangle}).

\newpage
\bibliographystyle{IEEEtranN}
\bibliography{sample}
\end{document}